\documentclass[10pt,twocolumn,letterpaper]{article}

\usepackage[pagenumbers]{cvpr} %

\usepackage{graphicx}
\usepackage{amsmath}
\usepackage{amssymb}
\usepackage{booktabs}
\usepackage[numbers,sort,compress]{natbib}

\usepackage{adjustbox}
\usepackage{multirow}
\usepackage[usenames,dvipsnames]{xcolor}
\usepackage{enumitem}

\usepackage[linesnumbered,ruled,vlined]{algorithm2e}
\usepackage{bbm}
\usepackage{amsfonts}       %
\usepackage{nicefrac}       %
\usepackage{mathtools}
\usepackage[accsupp]{axessibility} %

\SetKwInput{KwInput}{Input}                %
\SetKwInput{KwOutput}{Output}              %

\makeatletter
\def\NAT@def@citea{\def\@citea{\NAT@separator}}
\makeatother

\usepackage{microtype}      %
\usepackage[skip=4pt]{caption}

\usepackage[breaklinks,colorlinks,citecolor=ForestGreen]{hyperref}

\usepackage[capitalize]{cleveref}
\crefname{section}{Sec.}{Secs.}
\Crefname{section}{Section}{Sections}
\Crefname{table}{Table}{Tables}
\crefname{table}{Tab.}{Tabs.}

\def\cvprPaperID{107} %
\def\confName{CVPR}
\def\confYear{2022}

\begin{document}

\title{StyleT2I: Toward Compositional and High-Fidelity Text-to-Image Synthesis}

\author{Zhiheng Li$^{1,2}$, Martin Renqiang Min$^1$, Kai Li$^1$, and Chenliang Xu$^2$\\
$^1$NEC Laboratories America, $^2$University of Rochester\\
{\tt\small \{renqiang,kaili\}@nec-labs.com, \{zhiheng.li,chenliang.xu\}@rochester.edu}\\
}
\maketitle

\begin{abstract}
Although progress has been made for text-to-image synthesis, previous methods fall short of generalizing to unseen or underrepresented attribute compositions in the input text. Lacking compositionality could have severe implications for robustness and fairness, e.g., inability to synthesize the face images of underrepresented demographic groups. In this paper, we introduce a new framework, StyleT2I, to improve the compositionality of text-to-image synthesis. Specifically, we propose a CLIP-guided Contrastive Loss to better distinguish different compositions among different sentences. To further improve the compositionality, we design a novel Semantic Matching Loss and a Spatial Constraint to identify attributes' latent directions for intended spatial region manipulations, leading to better disentangled latent representations of attributes. Based on the identified latent directions of attributes, we propose Compositional Attribute Adjustment to adjust the latent code, resulting in better compositionality of image synthesis.
In addition, we leverage the $\ell_2$-norm regularization of identified latent directions (norm penalty) to strike a nice balance between image-text alignment and image fidelity. In the experiments, we devise a new dataset split and an evaluation metric to evaluate the compositionality of text-to-image synthesis models. The results show that StyleT2I outperforms previous approaches in terms of the consistency between the input text and synthesized images and achieves higher fidelity.
\end{abstract}

\begin{figure}[t]
  \centering
  \includegraphics[width=\linewidth]{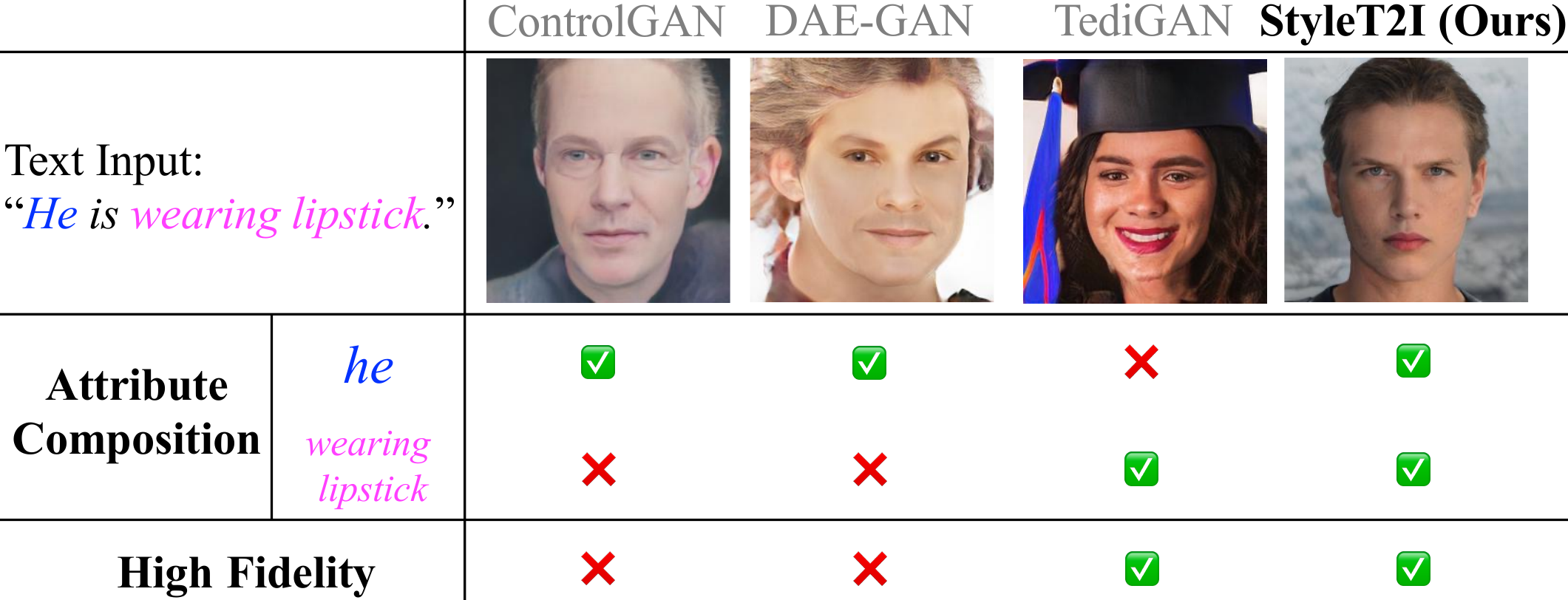}
  \caption{When the text input contains underrepresented compositions of attributes, \eg, (\textcolor{blue}{\textit{he}}, \textcolor{magenta}{\textit{wearing lipstick}}), in the dataset, previous methods~\cite{li2019Adv.NeuralInf.Process.Syst.b,ruan2021IEEEInt.Conf.Comput.Vis.ICCV,xia2021IEEEConf.Comput.Vis.PatternRecognit.CVPR} incorrectly generate the attributes with poor image quality. In contrast, StyleT2I achieves better compositionality and high-fidelity text-to-image synthesis results.
  }
   \label{fig.teaser}
\end{figure}

\section{Introduction}

Text-to-image synthesis is a task to synthesize an image conditioned on %
given input text, which enables many downstream applications, such as art creation,  computer-aided design, and training data generation for augmentation. Although progress has been made for this task, the compositionality aspect %
is overlooked by many previous methods~\cite{park2021Thirty-FifthConf.NeuralInf.Process.Syst.DatasetsBenchmarksTrackRound1}. As shown in \cref{fig.teaser}, the input text ``\textit{he\footnote{In this work, the gender and gender pronouns denote the visually perceived gender, which does not indicate one's actual gender identity.} is wearing lipstick}'' describes an intersectional group~\cite{buolamwini2018ACMConf.FairnessAccount.Transpar.} between two attributes---``\textit{he}'' and ``\textit{wearing lipstick},'' which is underrepresented in a face dataset~\cite{karras2018Int.Conf.Learn.Represent.}. The previous approaches~\cite{li2019Adv.NeuralInf.Process.Syst.b,ruan2021IEEEInt.Conf.Comput.Vis.ICCV,xia2021IEEEConf.Comput.Vis.PatternRecognit.CVPR} fail to correctly synthesize the image,
which could be caused by overfitting to the overrepresented compositions, \eg, (``\textit{she}'', ``\textit{wearing lipstick}'') and (``\textit{he}'', not ``\textit{wearing lipstick}''), in the dataset. %
This leads to severe robustness and fairness issues by inheriting biases and stereotypes from the dataset. Therefore, it is imperative to improve %
the text-to-image synthesis results in the aspect of compositionality.

The crux of the compositionality problem is to prevent %
models from simply memorizing the compositions in the training data. First, in terms of the objective function, some previous methods~\cite{xia2021IEEEConf.Comput.Vis.PatternRecognit.CVPR,xia2021ArXiv210408910Cs} simply minimize the feature distance %
between pairwise matched image and text, leading to poor generalizability. In contrast, we propose a \textit{CLIP-guided Contrastive Loss} to let the network better distinguish different compositions among different sentences, in which CLIP (Contrastive Language–Image Pre-training)~\cite{radford2021Int.Conf.Mach.Learn.} is pre-trained on large-scale matched image-text pairs as a foundation model~\cite{bommasani2021ArXiv210807258Cs}. Second, the compositional text-to-image model needs to be sensitive to each independent attribute described in the text. Most previous methods~\cite{xu2018IEEEConf.Comput.Vis.PatternRecognit.CVPRa,li2019Adv.NeuralInf.Process.Syst.b,zhu2019IEEEConf.Comput.Vis.PatternRecognit.CVPRc,zhang2021IEEEConf.Comput.Vis.PatternRecognit.CVPRa} mainly resort to attention mechanism~\cite{vaswani2017Adv.NeuralInf.Process.Syst.}, which focuses more on the correspondence between words and image features but falls short of separating individual attributes from a composition. Unlike previous approaches, our key idea is to identify disentangled representations~\cite{chen2016Adv.NeuralInf.Process.Syst.,higgins2017Int.Conf.Learn.Represent.} in the latent space of a generative model, where each disentangled representation exclusively corresponds to one attribute in the dataset. By leveraging the disentangled representations of different attributes, we can improve the compositionality by ensuring that each attribute described in the sentence is correctly synthesized.

Motivated by these ideas, we present StyleT2I, a %
novel framework to improve the compositionality of text-to-image synthesis %
employing StyleGAN~\cite{karras2019IEEEConf.Comput.Vis.PatternRecognit.CVPR}. In specific, we propose a \textit{CLIP-guided Contrastive Loss} to train a network to find the StyleGAN's latent code semantically aligned with the input text and better distinguish different compositions in different sentences. To further improve the compositionality, we propose a \textit{Semantic Matching Loss} and a \textit{Spatial Constraint} for identifying attributes' latent directions that induce intended spatial region manipulations. %
This leads to a better disentanglement of latent representations for different attributes. Then we propose \textit{Compositional Attribute Adjustment} to correct the wrong attribute synthesis by adjusting the latent code based on identified attribute directions during the inference stage. However, we empirically found that optimizing the proposed losses above can sometimes lead to degraded image quality. To address this issue, we %
employ \textit{norm penalty} to strike a nice balance between image-text alignment and image fidelity.

To better evaluate the compositionality of text-to-image synthesis, we devise a %
test split for the  CelebA-HQ~\cite{karras2018Int.Conf.Learn.Represent.} dataset, where the %
test text only contains unseen compositions of attributes. We design a new evaluation metric for the CUB~\cite{wah2011} dataset to evaluate if the synthesized image is in the correct bird species. %
Extensive quantitative results, qualitative results, and user studies manifest the advantages of our method on both image-text alignment and fidelity for compositional text-to-image synthesis.

We summarize our contributions as follows:
(1) We propose StyleT2I, a compositional text-to-image synthesis framework with a novel \textit{CLIP-guided Contrastive Loss} and \textit{Compositional Attribute Adjustment}. To the best of our knowledge, this is the first text-to-image synthesis work that focuses on improving the compositionality of different attributes.
(2) We propose a novel \textit{Semantic Matching Loss} and a \textit{Spatial Constraint} for identifying attributes' latent directions that induce intended variations in the image space, leading to a better disentanglement among different attributes.
(3) We devise a new test split and an evaluation metric to better evaluate the compositionality of text-to-image synthesis.

\section{Related Work}

\noindent \textbf{Text-to-Image Synthesis} Many previous works~\cite{reed2016Int.Conf.Mach.Learn.,zhang2017IEEEInt.Conf.Comput.Vis.ICCVa,zhang2019IEEETrans.PatternAnal.Mach.Intell.,xu2018IEEEConf.Comput.Vis.PatternRecognit.CVPRa,li2019Adv.NeuralInf.Process.Syst.b,zhang2021IEEEConf.Comput.Vis.PatternRecognit.CVPRa,zhu2019IEEEConf.Comput.Vis.PatternRecognit.CVPRc,yin2019IEEEConf.Comput.Vis.PatternRecognit.CVPR,qiao2019IEEEConf.Comput.Vis.PatternRecognit.CVPRa,cheng2020IEEEConf.Comput.Vis.PatternRecognit.CVPRd,ramesh2021Int.Conf.Mach.Learn.,vandenoord2017Adv.NeuralInf.Process.Syst.,liang2020Eur.Conf.Comput.Vis.ECCV,hinz2020IEEETrans.PatternAnal.Mach.Intell.,li2019IEEEConf.Comput.Vis.PatternRecognit.CVPRf,koh2021IEEEWinterConf.Appl.Comput.Vis.WACV} have %
studied text-to-image synthesis. %
DALL$\cdot$E~\cite{ramesh2021Int.Conf.Mach.Learn.} trains dVAE~\cite{vandenoord2017Adv.NeuralInf.Process.Syst.} that autoregressively predicts the image tokens on a large-scale dataset. \citet{zhang2021IEEEConf.Comput.Vis.PatternRecognit.CVPRa} use cross-modal contrastive loss on real image-text and fake image-real image pairs to adversarially train the conditional GAN. In contrast, StyleT2I's \textit{CLIP-guided Contrastive Loss} enjoys a simpler training scheme by using the pretrained CLIP as a conditional discriminator to contrast fake image-text pairs. While DAE-GAN~\cite{ruan2021IEEEInt.Conf.Comput.Vis.ICCV} extracts aspects from the language with the attention mechanism to improve image-text alignment, StyleT2I identifies attribute's latent directions and explicitly manipulates the latent code with proposed \textit{Compositional Attribute Adjustment}, which is more interpretable. TediGAN~\cite{xia2021IEEEConf.Comput.Vis.PatternRecognit.CVPR,xia2021ArXiv210408910Cs} uses pretrained StyleGAN~\cite{karras2019IEEEConf.Comput.Vis.PatternRecognit.CVPR} as the generator and trains a text encoder by deterministically minimizing the feature distances between paired image and text in either StyleGAN's latent space~\cite{xia2021IEEEConf.Comput.Vis.PatternRecognit.CVPR} or CLIP's feature space~\cite{xia2021ArXiv210408910Cs}, which suffers from memorizing the dataset's compositions. TediGAN also needs to conduct a manual analysis to find the layer-wise control for each attribute. In comparison, StyleT2I automatically finds disentangled latent directions for different attributes with a novel \textit{Semantic Matching Loss} and a \textit{Spatial Constraint}. \citet{wang2021IEEEWinterConf.Appl.Comput.Vis.WACVa} perform text-to-face synthesis based on attribute's latent direction identified by using additional attribute labels as supervision, whereas StyleT2I does not need additional attribute labels. \citet{tan2019IEEEConf.Comput.Vis.PatternRecognit.CVPR} focus on the compositionality problem for multi-object scene image synthesis.
Very recently, \citet{park2021Thirty-FifthConf.NeuralInf.Process.Syst.DatasetsBenchmarksTrackRound1} propose a new benchmark, revealing that many previous methods suffer from the compositionality problem, which motivates us to propose StyleT2I to address this issue.

\noindent \textbf{Disentangled Representation} %
Unsupervised disentangled representation learning focuses on training generative models~\cite{goodfellow2014Adv.NeuralInf.Process.Syst.,kingma2014Int.Conf.Learn.Represent.} %
with different latent dimensions interpreting independent factors of data variations, and most of such models are based on VAE~\cite{higgins2017Int.Conf.Learn.Represent.,kim2018Int.Conf.Mach.Learn.a,kingma2018Adv.NeuralInf.Process.Syst.,chen2018Adv.NeuralInf.Process.Syst.a,kumar2018Int.Conf.Learn.Represent.} and GAN~\cite{peebles2020Eur.Conf.Comput.Vis.ECCV,wei2021IEEEInt.Conf.Comput.Vis.ICCV}, enabling many downstream applications~\cite{lang2021IEEEInt.Conf.Comput.Vis.ICCV,li2021IEEEInt.Conf.Comput.Vis.ICCV,shi2022IEEEConf.Comput.Vis.PatternRecognit.CVPR}.  However, \citet{locatello2019Int.Conf.Mach.Learn.} show that unsupervised disentanglement is impossible without inductive bias or supervision. \citet{zhu2021IEEEConf.Comput.Vis.PatternRecognit.CVPR} modify the generative model's architecture with an additional loss to improve spatial constriction and variation simplicity. Some supervised disentanglement methods use a pre-trained classifier~\cite{shen2020IEEEConf.Comput.Vis.PatternRecognit.CVPR}, regressor~\cite{zhuang2021Int.Conf.Learn.Represent.}, or multi-attribute annotation~\cite{balakrishnan2020Eur.Conf.Comput.Vis.ECCV} as the full supervision to identify latent attribute directions. In contrast, StyleT2I finds disentangled attribute directions in the unmodified StyleGAN's latent space based on the supervision from text, which has a much lower annotation cost than multi-attribute labels.

\begin{figure}[t]
  \centering
   \includegraphics[width=\linewidth]{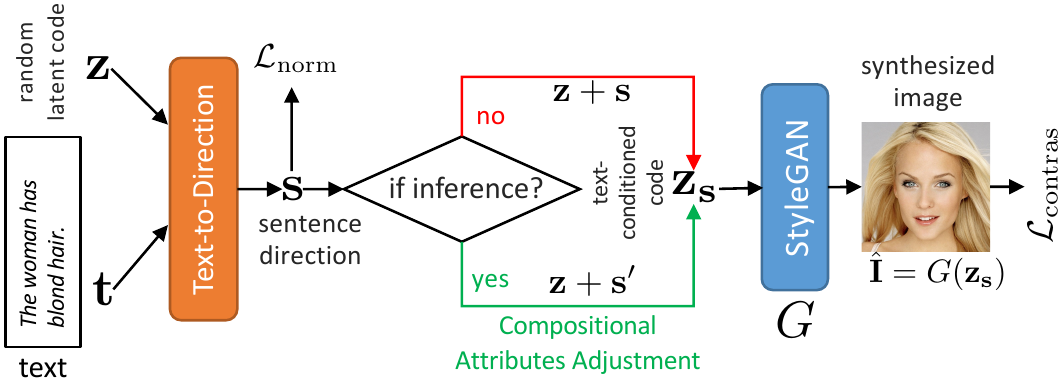}
   \caption{An overview of StyleT2I. The \textit{Text-to-Direction} module takes a text $\mathbf{t}$ and a random latent code $\mathbf{z}$ as inputs and outputs a sentence direction $\mathbf{s}$ to edit $\mathbf{z}$, resulting in a text-conditioned latent code $\mathbf{z}_\mathbf{s} = \mathbf{z} + \mathbf{s}$ in StyleGAN's latent space for image synthesis. The \textit{Text-to-Direction} module is trained with novel \textit{CLIP-guided Contrastive Loss} (\cref{subsec.clip_guided_contras_loss}) with \textit{norm penalty} employed (\cref{subsec.norm_penalty}). During the inference stage (lower branch), \textit{Compositional Attribute Adjustment} (\cref{subsec.compositional_attr_adjustment}) is performed by adjusting $\mathbf{s}$ to $\mathbf{s}'$, leading to better compositionality.}
   \label{fig.overview}
\end{figure}

\section{Overview of StyleT2I}
\Cref{fig.overview} gives an overview of our StyleT2I framework. Unlike most previous end-to-end approaches~\cite{xu2018IEEEConf.Comput.Vis.PatternRecognit.CVPRa,zhu2019IEEEConf.Comput.Vis.PatternRecognit.CVPRc,zhang2021IEEEConf.Comput.Vis.PatternRecognit.CVPRa,ruan2021IEEEInt.Conf.Comput.Vis.ICCV}, we leverage a pre-trained unconditional generator, StyleGAN~\cite{karras2019IEEEConf.Comput.Vis.PatternRecognit.CVPR}, and focus on finding a text-conditioned latent code in the generator's latent space that can be decoded into a high-fidelity image aligned with the input text.

To achieve this, in \cref{sec.text_to_latent}, we present a \textit{Text-to-Direction} module (see \cref{fig.overview}) trained with a novel \textit{CLIP-guided Contrastive Loss} for better distinguishing different compositions (\cref{subsec.clip_guided_contras_loss}) and %
a \textit{norm penalty} (\cref{subsec.norm_penalty}) to preserve the high fidelity of the synthesized image.

To further improve the compositionality of the text-to-image synthesis results, in \cref{sec.improve_compositionality_w_attr_dir}, we propose a novel \textit{Semantic Matching Loss} (\cref{subsec.semantic_matching_loss}) and a \textit{Spatial Constraint} (\cref{subsec.spatial_supervision}) for identifying disentangled attribute latent directions, which will be used to adjust the text-conditioned latent code during the inference stage (\cref{subsec.compositional_attr_adjustment}) with our novel \textit{Compositional Attribute Adjustment} (\textit{CAA}). The pseudocode of the complete algorithm is in \cref{subsec.supp.complete_algo}.

\section{Text-conditioned Latent Code Prediction}
\label{sec.text_to_latent}

As many previous works~\cite{shen2020IEEETrans.PatternAnal.Mach.Intell.,patashnik2021IEEEInt.Conf.Comput.Vis.ICCV,zhuang2021Int.Conf.Learn.Represent.,shen2020IEEEConf.Comput.Vis.PatternRecognit.CVPR,yao2021IEEEInt.Conf.Comput.Vis.ICCV} show that the latent direction in StyleGAN's latent space can represent an attribute---traversing a latent code along the attribute's latent direction can edit the attribute in the synthesized image, we hypothesize that there exists a latent direction that corresponds to the composition of multiple attributes described in the input text, \eg, ``\textit{woman}'' and ``\textit{blond hair}'' attributes in text ``\textit{the woman has blond hair}.'' Therefore, to find a text-conditioned latent code in a pre-trained StyleGAN's latent space, we propose a \textit{Text-to-Direction} module that takes the text $\mathbf{t}$ and a randomly sampled latent code $\mathbf{z}$ from the latent space of the pre-trained StyleGAN as input. The output is a latent direction $\mathbf{s}$, dubbed sentence direction, to edit the latent code $\mathbf{z}$, resulting in the text-conditioned code $\mathbf{z}_\mathbf{s} = \mathbf{z} + \mathbf{s}$. As a result, $\mathbf{z}_\mathbf{s}$ is fed into the StyleGAN generator $G$ to synthesize the image $\hat{\mathbf{I}} = G(\mathbf{z_s})$.

\subsection{CLIP-guided Contrastive Loss}
\label{subsec.clip_guided_contras_loss}
The \textit{Text-to-Direction} module should predict the sentence direction that is aligned with the input text and avoid simply memorizing the compositions in the training data. To achieve this, we leverage a foundational model  CLIP~\cite{sohn2016Adv.NeuralInf.Process.Syst.} pre-trained on a large-scale dataset with matched image-caption pairs to learn a joint embedding space of text and image, as a conditional discriminator. We propose a novel \textit{CLIP-guided Contrastive Loss} based on CLIP and contrastive loss~\cite{chen2020Int.Conf.Mach.Learn.b} to train the \textit{Text-to-Direction} module. Formally, given a batch of $B$ text $\{\mathbf{t}_i\}_{i=1}^B$ sampled from the training data and the corresponding fake images $\hat{\mathbf{I}}_i$, we compute the \textit{CLIP-guided Contrastive Loss} of the $i$-th fake image as:
\begin{equation}
    \mathcal{L}_\text{contras}(\mathbf{I}_i) = -\log \frac{\exp (\cos (E_\text{CLIP}^\text{img}(\mathbf{\hat{I}}_i), E_\text{CLIP}^\text{text}(\mathbf{t}_i) ) ) }{\sum_{j \neq i}^B \exp (\cos (E_\text{CLIP}^\text{img}(\mathbf{\hat{I}}_i), E_\text{CLIP}^\text{text}(\mathbf{t}_j) )) },
    \label{eq.loss_contras}
\end{equation}
where $E_\text{CLIP}^\text{img}$ and $E_\text{CLIP}^\text{text}$ denote the image encoder and text encoder of CLIP, respectively. $\cos(\cdot, \cdot)$ denotes the cosine similarity. \textit{CLIP-guided Contrastive Loss} attracts paired text embedding and fake image embedding in CLIP's joint feature space and repels the embedding of unmatched pairs. In this way, the \textit{Text-to-Direction} module is trained to better align the sentence direction $\mathbf{s}$ with the input text $\mathbf{t}$. At the same time, \textit{CLIP-guided Contrastive Loss} forces the \textit{Text-to-Direction} module to contrast the different compositions in different texts, \eg, ``\textit{he is wearing lipstick}'' and ``\textit{she is wearing lipstick},'' which prevents the network from overfitting to compositions that predominate in the training data.

\subsection{Norm Penalty for High-Fidelity Synthesis}
\label{subsec.norm_penalty}
However, the experimental results (\cref{fig.ablate_norm_penalty}) show that minimizing the contrastive loss alone fails to guarantee the fidelity of the synthesized image. We observe that it makes the \textit{Text-to-Direction} module predict $\mathbf{s}$ with a large $\ell_2$ norm, resulting in $\mathbf{z}_\mathbf{s}$ shifted to the low-density region in the latent distribution, leading to degraded image quality.
Therefore, we penalize the $\ell_2$ norm of sentence direction $\mathbf{s}$ when it exceeds a threshold hyperparameter $\theta$:
\begin{equation}
    \mathcal{L}_\text{norm} = \max (|| \mathbf{s} ||_2 - \theta, 0).
    \label{eq.norm_penalty}
\end{equation}
Our ablation study (\cref{fig.ablate_norm_penalty}) shows that adding the \textit{norm penalty} strikes a nice balance between the text-image alignment and quality.

To summarize, the \textbf{full objective function} for training the \textit{Text-to-Direction} module is:
\begin{equation}
    \mathcal{L}_s = \mathcal{L}_\text{contras} + \mathcal{L}_\text{norm}.
\end{equation}

\section{Compositionality with Attribute Directions}
\label{sec.improve_compositionality_w_attr_dir}

To further improve the compositionality, we first identify the latent directions representing the attributes with a novel \textit{Semantic Matching Loss} (\cref{subsec.semantic_matching_loss}) and a \textit{Spatial Constraint} (\cref{subsec.spatial_supervision}). Then, we propose \textit{Compositional Attribute Adjustment} (\cref{subsec.compositional_attr_adjustment}) to adjust the sentence direction by the identified attribute directions to improve the compositionality of text-to-image synthesis results.

\begin{figure}[t]
  \centering
    \includegraphics[width=\linewidth]{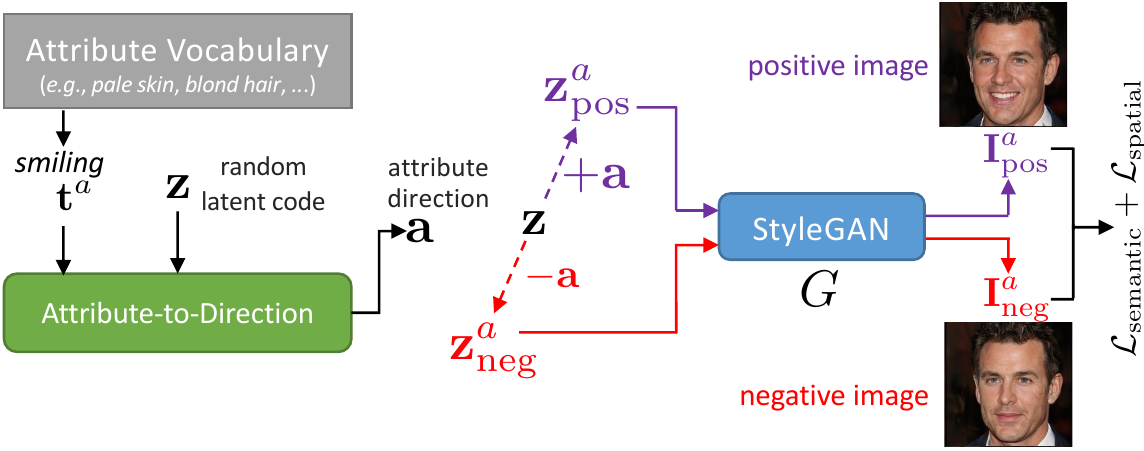}
   \caption{Identifying disentangled attribute directions by training an \textit{Attribute-to-Direction} module with a \textit{Semantic Matching Loss} ($\mathcal{L}_\text{semantic}$) and a \textit{Spatial Constraint} ($\mathcal{L}_\text{spatial}$).}
   \label{fig.triplet}
\end{figure}

\subsection{Identify Attribute Directions via a  Semantic Matching Loss}
\label{subsec.semantic_matching_loss}

To identify the latent directions of attributes existing in the dataset, we first build a vocabulary of attributes, \eg, ``\textit{smiling},'' ``\textit{blond hair},'' attributes in a face dataset, where each attribute is represented by a word or a short phrase. Then, we extract the attributes from each sentence in the dataset based on string matching or dependency parsing. For example, ``\textit{woman}'' and ``\textit{blond hair}'' attributes are extracted from the sentence ``\textit{the woman has blond hair}.''

Then, we present an \textit{Attribute-to-Direction} module (see \cref{fig.triplet}) that takes the random latent code $\mathbf{z}$ and word embedding of attributes $\mathbf{t}^a$ sampled from the attribute vocabulary as the inputs, outputting the attribute direction $\mathbf{a}$. To ensure that $\mathbf{a}$ is semantically matched with the input attribute, we propose a novel \textit{Semantic Matching Loss} to train the \textit{Attribute-to-Direction} module. Concretely, $\mathbf{a}$ is used to edit $\mathbf{z}$ to obtain the positive latent code $\mathbf{z}^a_\text{pos} = \mathbf{z} + \mathbf{a}$ and negative latent code $\mathbf{z}^a_\text{neg} = \mathbf{z} - \mathbf{a}$. $\mathbf{z}^a_\text{pos}$ is used to synthesize the positive image $\mathbf{I}^a_\text{pos} = G(\mathbf{z}^a_\text{pos})$ that can reflect the semantic meaning of the attribute, \eg, the smiling face in \cref{fig.triplet}. While $\mathbf{z}^a_\text{neg} = G(\mathbf{z}^a_\text{neg})$ is used to synthesize the negative image $\mathbf{I}^a_\text{neg} = G(\mathbf{z}^a_\text{neg})$ that does \textit{not} contain the information of the given attribute, \eg, the \textit{not} smiling face in \cref{fig.triplet}. Based on the triplet~\cite{schroff2015IEEEConf.Comput.Vis.PatternRecognit.CVPR} of ($\mathbf{t}^a$, $\mathbf{I}^a_\text{pos}$, $\mathbf{I}^a_\text{neg}$), the \textit{Semantic Matching Loss} is computed as:
\begin{equation}
\begin{aligned}
    \mathcal{L}_\text{semantic} = & \max( \cos (E_\text{CLIP}^\text{img}(\mathbf{I}^a_\text{neg}), E_\text{CLIP}^\text{text}(\mathbf{t}^a))  \\
    &- \cos (E_\text{CLIP}^\text{img}(\mathbf{I}^a_\text{pos}), E_\text{CLIP}^\text{text}(\mathbf{t}^a)) + \alpha , 0),
    \label{eq.semantic_match_triplet}
\end{aligned}
\end{equation}
where $\alpha$ is a hyperparameter as the margin. $\mathcal{L}_\text{semantic}$ attracts attribute text embedding and positive image embedding and repels the attribute text embedding against negative image embedding in CLIP's feature space, rendering the attribute direction $\mathbf{a}$ semantically matched with the attribute.

\begin{figure}[t]
  \centering
    \includegraphics[width=\linewidth]{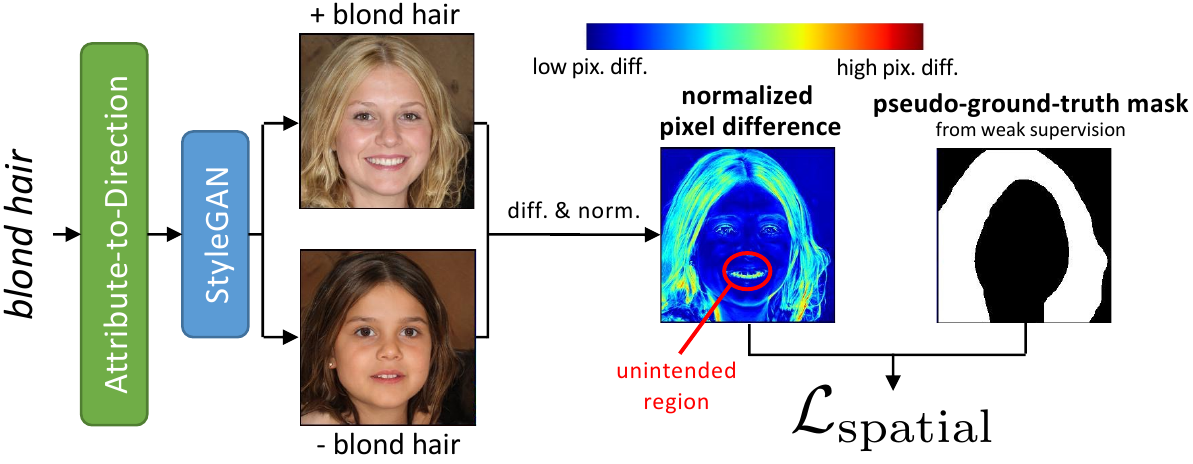}
   \caption{\textit{Spatial Constraint} ($\mathcal{L}_\text{spatial}$) to train \textit{Attribute-to-Direction} module. We compute the pixel-level difference between the positive and negative image to measure the changing region on the image space (red: high pixel differences; blue: low pixel variations). $\mathcal{L}_\text{spatial}$ supervises the pixel-level differences by the mask (obtained from a weak-supervised segmentation method) of the intended region (\eg, hair) for the given attribute (\eg, ``\textit{blond hair}'') to suppress changes on other unintended areas (\eg, mouth), leading to better disentanglement among different attributes.}
   \label{fig.spatial_reg}
\end{figure}

\subsection{Attribute Disentanglement with a Spatial Constraint}
\label{subsec.spatial_supervision}

However, the \textit{Semantic Matching Loss} cannot ensure that the given attribute is disentangled with other attributes. For example, in \cref{fig.spatial_reg}, while the \textit{Attribute-to-Direction} module is expected to predict an attribute direction of ``\textit{blond hair},'' the mouth region is also changing. To mitigate this issue, we propose a novel \textit{Spatial Constraint} as an additional loss to train the \textit{Attribute-to-Direction} module. Our motivation is to restrict the spatial variation between the positive and negative images to an intended region, \eg, the hair region for the ``\textit{blond hair}'' attribute. To achieve this, we capture the spatial variation by computing the pixel-level difference $\mathbf{I}_\text{diff}^a = \sum_c |\mathbf{I}_\text{pos}^a - \mathbf{I}_\text{neg}^a|$, where $c$ denotes image channel dimension. Then, min-max normalization is applied to rescale its range to 0 to 1, denoted as $\mathbf{\tilde{I}}_\text{diff}^a$. We send the positive image to a weakly-supervised (\ie, supervised by attributes extracted from text) part segmentation method~\cite{huang2020IEEEConf.Comput.Vis.PatternRecognit.CVPRb} to acquire the pseudo-ground-truth mask $\mathbf{M}^a$ (\cref{subsec.implementation_details}), \eg, hair region mask in \cref{fig.spatial_reg}. Finally, \textit{Spatial Constraint} is computed as:
\begin{equation}
    \mathcal{L}_\text{spatial} = \text{BCE}(\mathbf{\tilde{I}}_\text{diff}^a, \mathbf{M}^a),
\end{equation}
where BCE denotes binary cross-entropy loss. Minimizing $\mathcal{L}_\text{spatial}$ will penalize the spatial variations out of the pseudo-ground-truth mask. In this way, the \textit{Attribute-to-Direction} module is forced to predict the attribute direction that can edit the image in the intended region.

In addition, similar to the \textit{norm penalty} used for \textit{Text-to-Direction} module, we also add it here to ensure the image quality. As a summary, the \textbf{full objective function} for training the \textit{Attribute-to-Direction} module is:
\begin{equation}
    \mathcal{L}_a = \mathcal{L}_\text{semantic} + \mathcal{L}_\text{spatial} + \mathcal{L}_\text{norm}.
\end{equation}

\subsection{Compositional Attribute Adjustment}
\label{subsec.compositional_attr_adjustment}

After training the \textit{Attribute-to-Direction} module, we propose novel \textit{Compositional Attribute Adjustment} (\textit{CAA}) to ensure the compositionality of the text-to-image synthesis results. The key idea of \textit{Compositional Attribute Adjustment} is two-fold. First, we identify the attributes that the sentence direction $\mathbf{s}$ incorrectly predicts based on its agreement with attribute directions. Second, once we identify the wrongly predicted attributes, we add these attribute directions as the correction to adjust the sentence direction.

Concretely, during the inference stage, as described in \cref{sec.text_to_latent}, we first sample a random latent code $\mathbf{z}$ and send it to \textit{Text-to-Direction} module along with the input text $\mathbf{t}$ to obtain the sentence direction $\mathbf{s}$. At the same time, we also extract $K$ attributes $\{\mathbf{t}^a_i\}_{i=1}^{K}$ from the sentence $\mathbf{t}$ and then feed it into the \textit{Attribute-to-Direction} module along with the random latent code $\mathbf{z}$ to obtain the attribute directions $\{\mathbf{a}_i\}_{i=1}^{K}$. Here $K$ is not a hyperparameter but is decided by the number of attributes described in the sentence, and the same $\mathbf{z}$ is used as the input for both the \textit{Text-to-Direction} module and the \textit{Attribute-to-Direction} module. Based on the attribute directions, we adjust the sentence direction $\mathbf{s}$ to $\mathbf{s}'$:
\begin{align}
    \mathbf{A} &= \{ \mathbf{a}_i \mid \cos(\mathbf{a}_i, \mathbf{s}) \leq 0 \}, &  \mathbf{s}' &= \mathbf{s} + \sum_{\mathbf{a}_i \in \mathbf{A}} \frac{\mathbf{a}_i}{||\mathbf{a}_i||_2},
\end{align}
where $\cos(\cdot, \cdot)$ denotes cosine similarity and $\mathbf{s}'$ stands for the attribute-adjusted sentence direction. $\mathbf{A}$ is a set of attribute directions that have a less or equal to zero cosine similarity with the sentence direction. When $\cos(\mathbf{a}_i, \mathbf{s}) \leq 0$, the sentence direction $\mathbf{s}$ is not agreed with the $i$-th attribute direction $\mathbf{a}_i$, indicating that $\mathbf{s}$ fails to reflect the $i$-th attribute in the input text. By adding the $i$-th attribute direction $\frac{\mathbf{a}_i}{||\mathbf{a}_i||_2}$, the adjusted sentence direction $\mathbf{s}'$ is corrected to reflect the $i$-th attribute. Then, it replaces $\mathbf{s}$ to edit the latent code $\mathbf{z}$ to obtain the new text-conditioned code $\mathbf{z}_\mathbf{s} = \mathbf{z} + \mathbf{s}'$ (lower branch in \cref{fig.overview}), which is used to synthesize the final image, enhancing compositionality of the text-to-image synthesis.

\section{Experiments}

\subsection{Experiment Setup}
\label{subsec.exp_setup}

\noindent \textbf{Dataset}
We use two datasets to conduct the experiments. The first dataset is CelebA-HQ~\cite{karras2018Int.Conf.Learn.Represent.}, which contains 30,000 celebrity face images. We use the text annotations provided by \citet{xia2021IEEEConf.Comput.Vis.PatternRecognit.CVPR}, where each text description is based on the facial attributes, \eg, ``\textit{She is wearing lipstick.}'' We remove the texts that mention the ``\textit{attractiveness}'' attribute due to the ethical concern~\cite{prabhu2019Int.Conf.Mach.Learn.Workshop}. The second dataset is CUB~\cite{wah2011}, which contains 11,788 bird images in 200 bird species. We use the text annotations collected by \citet{reed2016IEEEConf.Comput.Vis.PatternRecognit.CVPR}, where each sentence describes the fine-grained attributes of the bird.

\noindent \textbf{Test Split for Compositionality Evaluation} To better evaluate the compositionality of the text-to-image synthesis results, we carefully choose the test split on each dataset. We observe that about half of the texts in the standard test split~\cite{lee2020IEEEConf.Comput.Vis.PatternRecognit.CVPR} of CelebA-HQ dataset contain compositions of attributes seen in the training split. Therefore, we exclude these texts with seen compositions from the test split. As a result, the texts in the new test split only contain the unseen compositions of attributes, which can better evaluate the compositionality results.
Proposed Split (PS)~\cite{xian2017IEEEConf.Comput.Vis.PatternRecognit.CVPR,xian2019IEEETrans.PatternAnal.Mach.Intell.} is a CUB dataset split to benchmark the compositional zero-shot learning by splitting the dataset based on bird species. We choose the ``unseen test'' in PS as the test split, which can evaluate the model's capability of synthesizing images in 50 unseen bird categories.

\noindent \textbf{Evaluation Metrics}

\noindent \textit{FID}. We use FID~\cite{heusel2017Adv.NeuralInf.Process.Syst.} to evaluate image quality results. Lower values indicate better image quality.

\noindent \textit{R-Precision}. We use R-Precision~\cite{xu2018IEEEConf.Comput.Vis.PatternRecognit.CVPRa} that evaluates the top-1 retrieval accuracy as the major evaluation metric in image-text alignment.
We follow \cite{park2021Thirty-FifthConf.NeuralInf.Process.Syst.DatasetsBenchmarksTrackRound1} to use the CLIP finetuned on the whole dataset (including the test split) to compute the R-Precision results, which has been shown to be more aligned with human evaluation results. Higher R-Precision values indicate better alignment between text and image.

\noindent \textit{Bird Species Classification Accuracy}. As the models are expected to synthesize birds in unseen species on CUB dataset, we regard that a model that can more accurately synthesize birds in unseen bird species has better compositionality for disentangling different attributes from seen bird species. To this end, we propose a new evaluation metric---bird species classification accuracy for evaluating compositionality. Concretely, we finetune a ResNet-18~\cite{he2016IEEEConf.Comput.Vis.PatternRecognit.CVPR} on the test split of CUB dataset with real images and bird species labels to classify 50 bird species. In evaluation, the test split contains (text, bird species label) pairs, where text is used to synthesize images. We use the finetuned classifier to predict bird species of the synthesized image. We report the top-1 accuracy based on the prediction and bird species labels (\cref{tab.cub_bird_cls}). However, a text may not contain enough discriminative information for classifying the bird species. Therefore, we train a text classifier, implemented as a GRU followed by an MLP, (last row in \cref{tab.cub_bird_cls}) that directly takes the text as input to predict the bird species. We train this text classifier on 80\% of texts in the test split, and we evaluate its classification accuracy on the rest 20\%, which can serve as the upper bound for the text-conditioned bird species classification results.

\noindent \textit{User Study}. The quantitative evaluation metrics above cannot substitute human evaluation. Therefore, we invite 12 subjects to conduct the user study on the two datasets to evaluate image quality and text alignment. Following \cite{zhang2021IEEEConf.Comput.Vis.PatternRecognit.CVPRa}, each question contains synthesized images from different methods conditioned on the same text input. Participants are invited to rank the synthesized images from different methods based on the image quality and image-text alignment. More details of the user study, \eg, user interface and use of human subjects, are in \cref{sec.supp.user_study}.

\noindent \textbf{Comparison Methods} We compare with four recent text-to-image synthesis methods---ControlGAN~\cite{li2019Adv.NeuralInf.Process.Syst.b}, DAE-GAN~\cite{ruan2021IEEEInt.Conf.Comput.Vis.ICCV}, TediGAN-A~\cite{xia2021IEEEConf.Comput.Vis.PatternRecognit.CVPR} TediGAN-B~\cite{xia2021ArXiv210408910Cs}. ControlGAN focuses on controllable generation based on attention mechanism. DAE-GAN extracts ``aspects'' information from text, which is related to the attributes studied in this paper. TediGAN-A trains a text encoder to minimize the distance between encoded text and encoded image in StyleGAN's latent space. TediGAN-B uses CLIP to optimize the StyleGAN’s latent code iteratively for each input text. For a fair comparison, we use the official code of each comparison method to conduct the experiments.

\begin{table}[t]
\centering
\begin{adjustbox}{width=\linewidth}
    \begin{tabular}{@{}l|ll|ll@{}}
    \toprule
                & \multicolumn{2}{c|}{CelebA-HQ}  & \multicolumn{2}{c}{CUB}         \\
                 & R-Precision $\uparrow$     & FID $\downarrow$           & R-Precision $\uparrow$     & FID $\downarrow$           \\ \midrule
    ControlGAN         & 0.435          & 31.38          & 0.137          & 29.03          \\
    DAE-GAN            & 0.484          & 30.74          & 0.145          & 26.99          \\
    TediGAN-A          & 0.044          & 16.45          & 0.071          & \textbf{16.38} \\
    TediGAN-B          & 0.306          & \textbf{15.46} & 0.121          & 16.79          \\
    StyleT2I (\textbf{Ours})    & \textbf{0.625} & 17.46          & \textbf{0.264} & 20.53          \\ \midrule
    StyleT2I-XD (\textbf{Ours}) & \textbf{0.698} & 18.02          & \textbf{0.350} & 19.19          \\ \bottomrule
    \end{tabular}
\end{adjustbox}
\caption{Text-to-Image synthesis results on CelebA-HQ~\cite{xia2021IEEEConf.Comput.Vis.PatternRecognit.CVPR} and CUB~\cite{wah2011} datasets. $\uparrow$: high values mean better results. $\downarrow$: lower values indicate better results.}
\label{tab.main_t2i}
\vspace{-5mm}
\end{table}

\subsection{Implementation Details}
\label{subsec.implementation_details}

\noindent \textbf{Architecture and Hyperparameters}
We choose StyleGAN2~\cite{karras2020IEEEConf.Comput.Vis.PatternRecognit.CVPR} as the generator for synthesizing images in $256^2$ resolution.
We use $\mathcal{W}+$ space as the latent space, where latent directions are more disentangled than the input noise space~\cite{karras2019IEEEConf.Comput.Vis.PatternRecognit.CVPR}.
GloVe~\cite{pennington2014Empir.MethodsNat.Lang.Process.} is used to obtain the word embeddings of text, which will be used as the input to \textit{Text-to-Direction} and \textit{Attribute-to-Direction} modules. The two modules have the same architecture---a GRU~\cite{chung2014Adv.NeuralInf.Process.Syst.Workshop} to extract the text feature, which is concatenated with the random latent code to send to a multi-layer perceptron with two fully-connected layers and one ReLU activation function~\cite{nair2010Int.Conf.Mach.Learn.}. We set the value $\theta=8$ in \cref{eq.norm_penalty} and $\alpha=1$ in \cref{eq.semantic_match_triplet}.
More details are in \cref{subsec.supp.hyperparameters_network_arch}.
The code is written in PyTorch~\cite{paszke2019Adv.NeuralInf.Process.Syst.} and is available at \url{https://github.com/zhihengli-UR/StyleT2I}.

\noindent \textbf{Attributes Vocabulary and Attributes Extraction}
For the vocabulary of attributes (\cref{subsec.semantic_matching_loss}), we use the attributes defined in ~\cite{liu2015IEEEInt.Conf.Comput.Vis.ICCV} (\eg, ``\textit{wearing lipstick}'') as the attributes of CelebA-HQ dataset, and the attributes defined in \cite{wah2011} (\eg, ``\textit{red belly}'') as the attributes of CUB dataset. Note that we do not use any attribute annotations. To extract attributes from sentences, we use string matching (\ie, the word ``lipstick'' in the sentence indicates ``\textit{wearing lipstick}'' attribute) on CelebA-HQ dataset. We use part-of-speech tag and dependency parsing implemented in spaCy~\cite{honnibal2017} to extract attributes from the text on CUB dataset. More details are shown in \cref{subsec.attr_extraction}.

\noindent \textbf{Pseudo-Ground-Truth Mask}
For the \textit{Spatial Constraint} (\cref{subsec.spatial_supervision}), we obtain the pseudo-ground-truth mask based on a weakly-supervised part segmentation method~\cite{huang2020IEEEConf.Comput.Vis.PatternRecognit.CVPRb}, where we train image classier supervised by attributes extracted from text.
More details are presented in \cref{subsec.supp.pseudo_gt_mask}.

\noindent \textbf{Finetune CLIP}
We empirically find that directly using the CLIP trained on the original large-scale dataset~\cite{radford2021Int.Conf.Mach.Learn.} performs poorly for the proposed losses (\cref{eq.loss_contras,eq.semantic_match_triplet}) on two datasets. We suspect the reason is the domain gap between in-the-wild images in the large-scale dataset~\cite{radford2021Int.Conf.Mach.Learn.} and face or birds images with fine-grained attributes. Therefore, we finetune the last few layers of CLIP on the training splits of CelebA-HQ and CUB datasets, respectively. Note that the CLIP used for training differs from the one used for evaluating R-Precision, where the latter is trained on the whole dataset.
More details are in \cref{subsec.supp.finetune_clip}.

\begin{table}[t]
  \centering
\begin{adjustbox}{width=0.65\linewidth}
      \begin{tabular}{@{}lcccc@{}}
  \toprule
  \multicolumn{1}{c}{Method}                  & Accuracy $\uparrow$       \\ \midrule
  ControlGAN               & 0.071          \\
  DAE-GAN                  & 0.056          \\
  TediGAN-A                  & 0.063          \\
  TediGAN-B   & 0.036 \\
  StyleT2I w/o (\textit{CAA})  (\textbf{Ours})   & 0.115          \\
  StyleT2I (\textbf{Ours})   & \textbf{0.125} \\ \midrule
  StyleT2I-XD (\textbf{Ours})   & \textbf{0.142} \\ \midrule
  Text Classifier (upper bound) & 0.204 \\ \bottomrule
  \end{tabular}
\end{adjustbox}
  \caption{Unseen bird species classification results. Our method outperforms other methods, and the results are closer to the upper bound, which demonstrates that StyleT2I can better synthesize unseen bird species based on the input text description, indicating better compositionality of our method.}
  \label{tab.cub_bird_cls}
  \vspace{-3mm}
\end{table}

\begin{figure*}[t]
  \centering
    \includegraphics[width=0.75\linewidth]{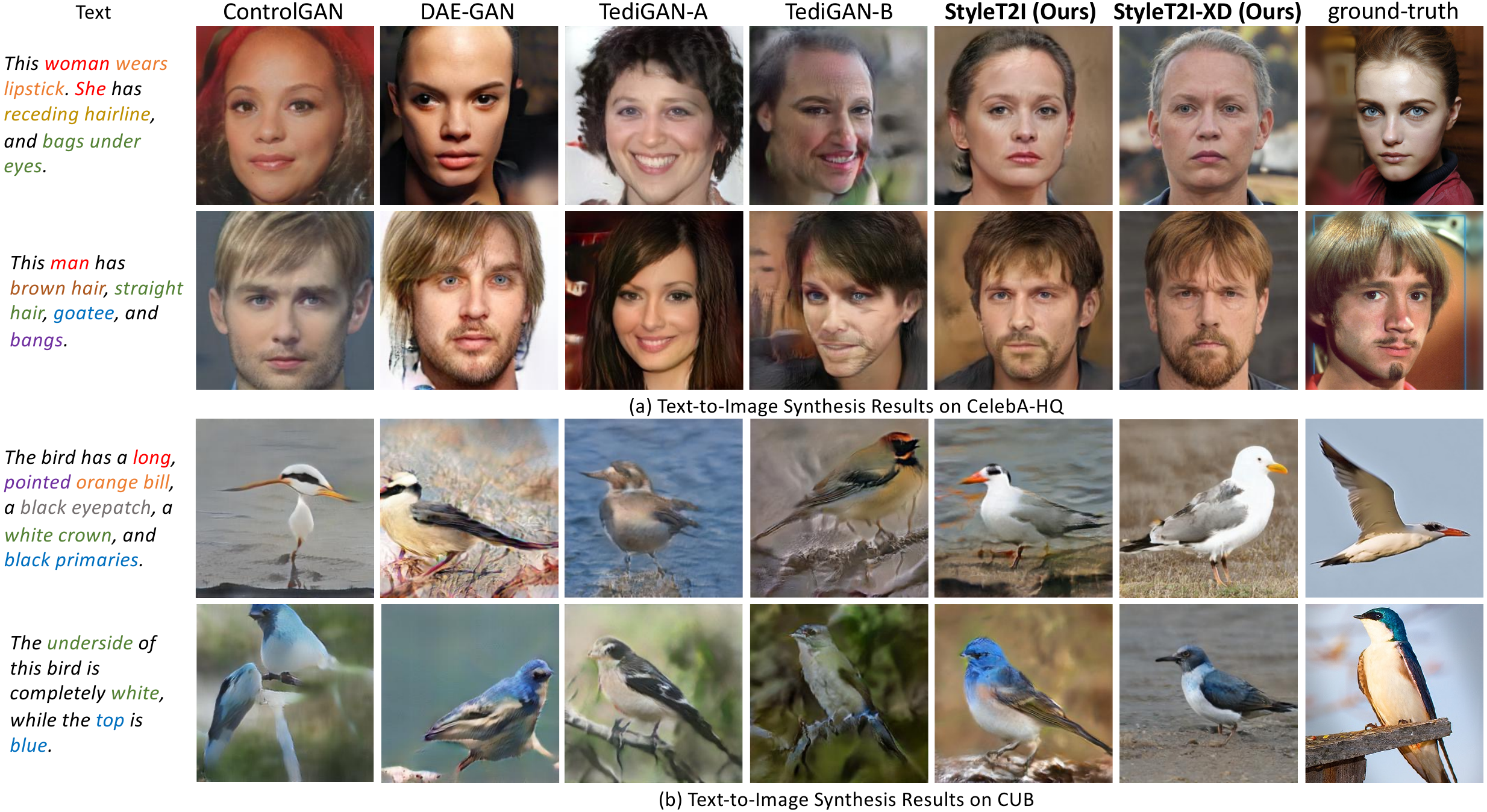}
  \caption{Qualitative comparison of text-to-image synthesis on CelebA-HQ and CUB datasets. Different attributes in the text are highlighted in different colors. More examples are in \cref{sec.supp.more_qualitative}.}
  \label{fig.t2i_comparison}
  \vspace{-3mm}
\end{figure*}

\noindent \textbf{Cross-dataset Synthesis (StyleT2I-XD)}
Since StyleT2I is based on a pretrained StyleGAN generator, we can train the StyleGAN generator on a different image dataset with more image samples and diversity to further improve the results. We denote this method as \textbf{StyleT2I-XD}. Concretely, we pretrain StyleGAN on FFHQ~\cite{karras2019IEEEConf.Comput.Vis.PatternRecognit.CVPR} dataset, a face dataset with more variation on various attributes (\eg, age), to synthesize images conditioned on the text from CelebA-HQ dataset. Similarly, we pretrain StyleGAN on NABirds~\cite{vanhorn2015IEEEConf.Comput.Vis.PatternRecognit.CVPR} dataset with more bird species (the unseen bird species in the test split are still excluded) and image samples to synthesize images conditioned on the text from CUB dataset.

\begin{figure}[t]
  \centering
    \includegraphics[width=\linewidth]{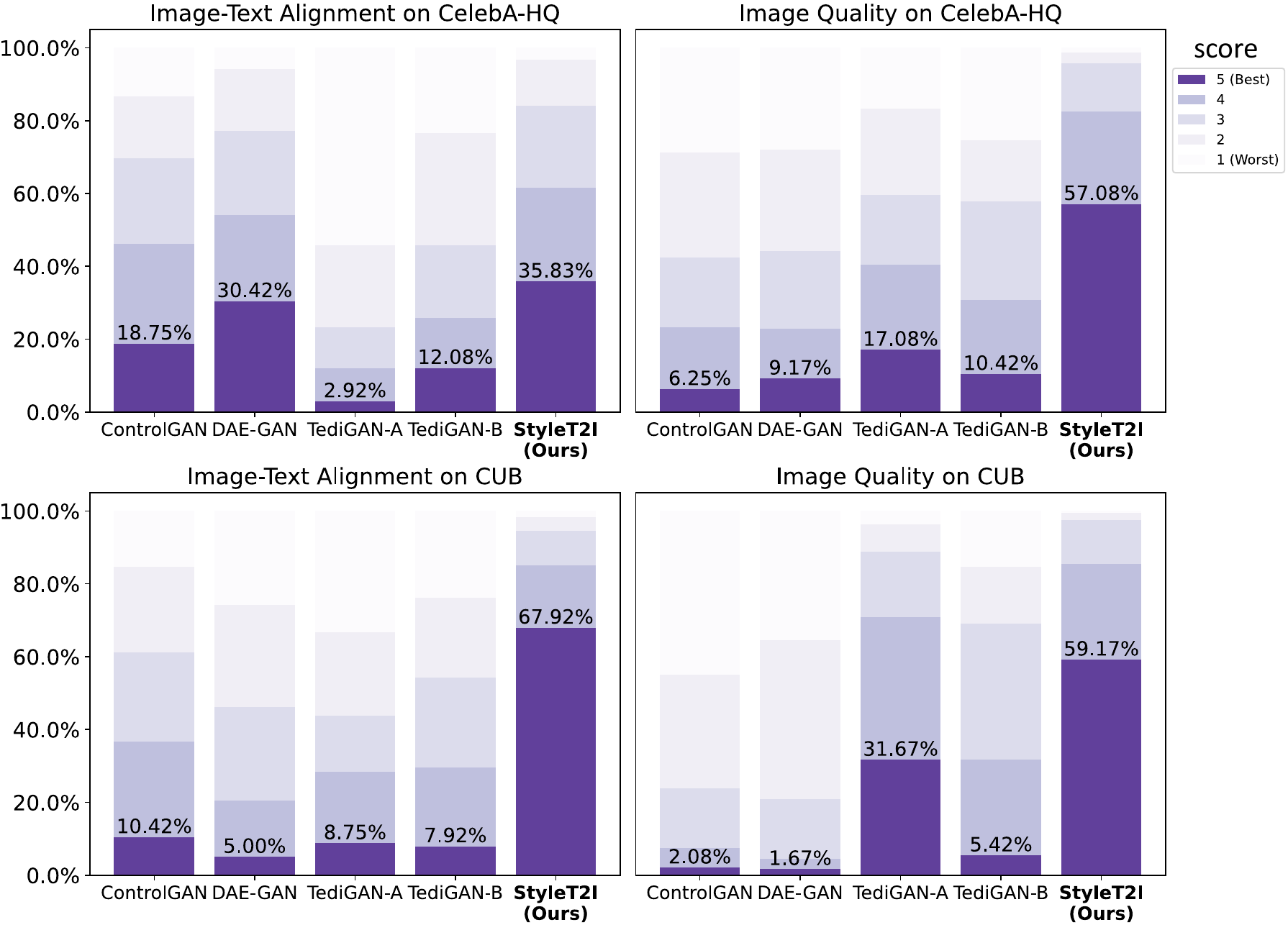}
   \caption{User study results on CelebA-HQ and CUB datasets.}
   \label{fig.user_study}
  \vspace{-5mm}
\end{figure}

\subsection{Results on Text-to-Image Synthesis}

\noindent \textbf{Quantitative Results} The quantitative results of text-to-image synthesis on CelebA-HQ and CUB datasets are shown in \cref{tab.main_t2i}. In terms of R-Precision, our StyleT2I outperforms other comparison methods by a large margin, showing that our method has a better compositionality to synthesize faces in novel compositions and birds in novel bird species. Although TediGAN-A is also based on StyleGAN, it performs poorly on both datasets, which suggests that deterministically minimizing the distance between the latent codes of text and image in StyleGAN's latent space leads to poor generalizability to the unseen compositions. The bird species classification results are shown in \cref{tab.cub_bird_cls}. Our StyleT2I outperforms other methods in the accuracy results by a large margin, which is also closer to the text classifier accuracy upper bound. This indicates that StyleT2I can more accurately synthesize the unseen bird species based on the text description, demonstrating better compositionality of StyleT2I. Concerning FID, our method achieves strong image quality results, which are also comparable with TediGAN. The FID results also show the advantage of StyleGAN-based methods (TediGAN and our StyleT2I) over methods with customized generator architectures (\ie, ControlGAN and DAE-GAN) for achieving high-fidelity synthesis results.

\noindent \textbf{Qualitative Results} We also show qualitative results in \cref{fig.t2i_comparison}. ControlGAN and DAE-GAN, although they reflect most attributes in the text, achieve poor images quality results. For example, in the first row of \cref{fig.t2i_comparison}, they both exaggerate the ``\textit{receding hairline}'' as bald. Although TediGAN can synthesize high-quality images, the images are barely aligned with the text, \eg, wrong gender in the second row of \cref{fig.t2i_comparison}. In contrast, the synthesized images by StyleT2I are in high fidelity and aligned with the attributes in text, \eg, ``\textit{orange bill}'' in \cref{fig.t2i_comparison} (b).

\noindent \textbf{User Study} The user study results are shown in \cref{fig.user_study}. Compared with other methods, StyleT2I receives higher ranking scores from the human participants in terms of both image-text alignment and image quality, which further manifests the advantages of our method.

\noindent \textbf{Cross-dataset Synthesis} Our cross-dataset text-to-image synthesis (StyleT2I-XD) can further improve the results. The quantitative results are shown in \cref{tab.main_t2i,tab.cub_bird_cls}. StyleT2I-XD achieves even stronger R-Precision and bird species classification accuracy results, demonstrating the effectiveness of cross-dataset training. Although StyleT2I-XD does not improve FID values, our qualitative results in \cref{fig.t2i_comparison} show that StyleT2I-XD achieves photo-realistic image quality.

\begin{table}[t]
\centering
\begin{adjustbox}{width=\linewidth}
  \begin{tabular}{@{}lcc@{}}
\toprule
                    & R-Precision $\uparrow$    & FID $\downarrow$           \\ \midrule
w/o \textit{CLIP-guided Contrastive Loss} &  0.205       & \textbf{18.64} \\
w/o \textit{norm penalty}    & \textbf{0.333} & \underline{23.86}    \\
w/o \textit{Spatial Constraint}   & 0.246          & \textbf{19.17} \\
w/o \textit{Compositional Attribute Adjustment} & 0.238          & \textbf{19.17} \\
w/o finetune CLIP         & \underline{0.145}     & 19.91          \\
Full Model          & \textbf{0.264} & 19.19 \\ \bottomrule
\end{tabular}
\end{adjustbox}
\caption{Ablation Study of StyleT2I on CUB dataset. Top-2 results are bolded and the worst results are underlined.}
\label{tab.ablate_t2i}
\end{table}

\begin{figure}[t]
  \centering
  \includegraphics[width=\linewidth]{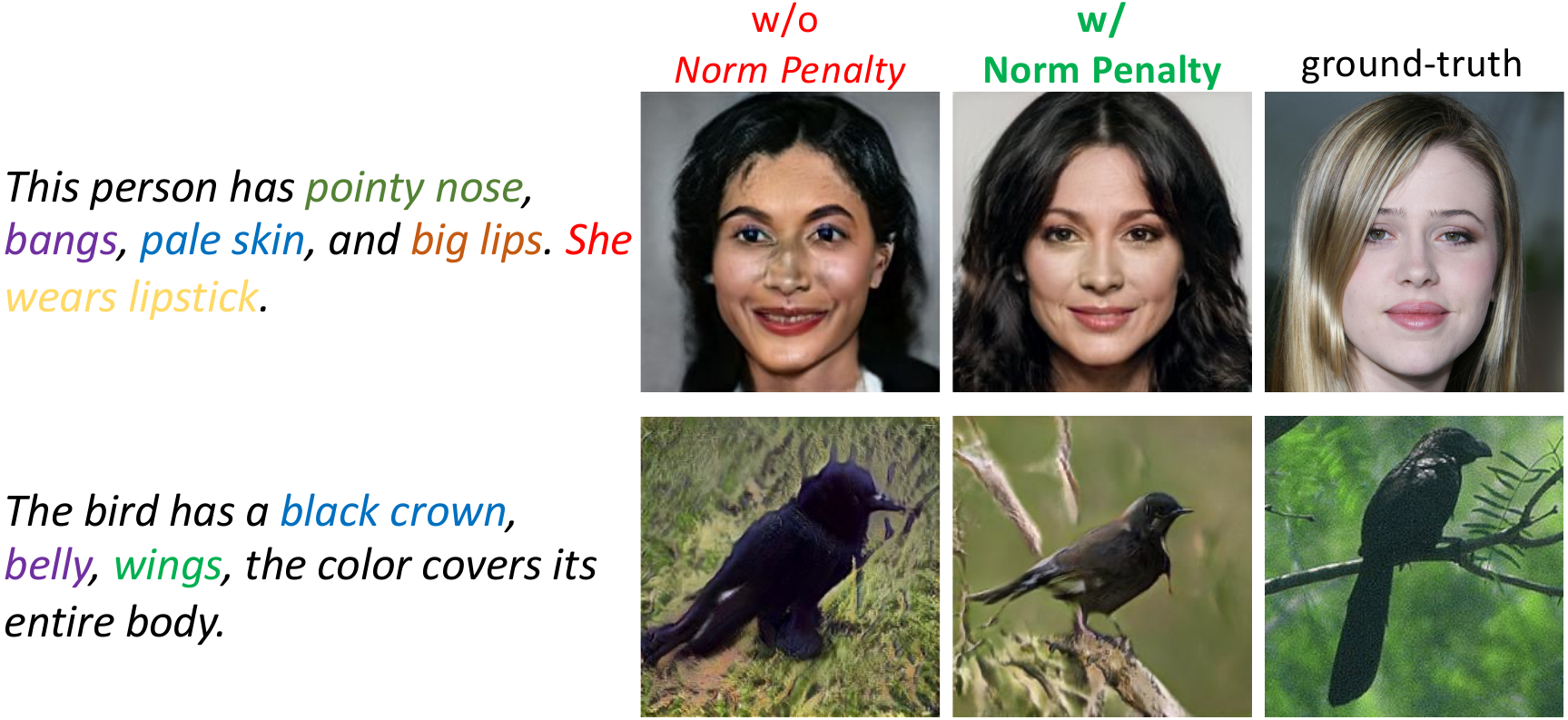}
  \caption{Ablation study of \textit{norm penalty} for improving image quality. More examples are shown in \cref{sec.supp.more_qualitative}.}
  \label{fig.ablate_norm_penalty}
  \vspace{-4mm}
\end{figure}

\subsection{Ablation Studies}
\label{subsec.ablation_study}

We conduct ablation studies to verify the effectiveness of each component of our method. More ablation study results are included in \cref{sec.supp.ablate_t2i,sec.supp.ablate_w2d}.

\noindent \textbf{CLIP-guided Contrastive Loss} An alternative loss to \cref{eq.loss_contras} is minimizing the cosine distance between the paired fake image feature and text feature in CLIP's feature space, which is initially proposed in StyleCLIP~\cite{patashnik2021IEEEInt.Conf.Comput.Vis.ICCV} and used in TediGAN-B~\cite{xia2021ArXiv210408910Cs} for text-to-image synthesis. The result of this alternative loss is shown on the first row of \cref{tab.ablate_t2i}. Although it slightly improves the FID result, the R-Precision result significantly decreases, demonstrating the necessity of contrasting unmatched (image, text) pairs to distinguish the difference of compositions better.

\noindent \textbf{Norm Penalty} As shown in \cref{tab.ablate_t2i} and \cref{fig.ablate_norm_penalty}, Although it lowers the performance in terms of R-Precision, using the proposed \textit{norm penalty} can effectively improve the FID results and perceptual quality, striking a better balance between image-text alignment and fidelity.

\begin{figure}[t]
  \centering
  \includegraphics[width=\linewidth]{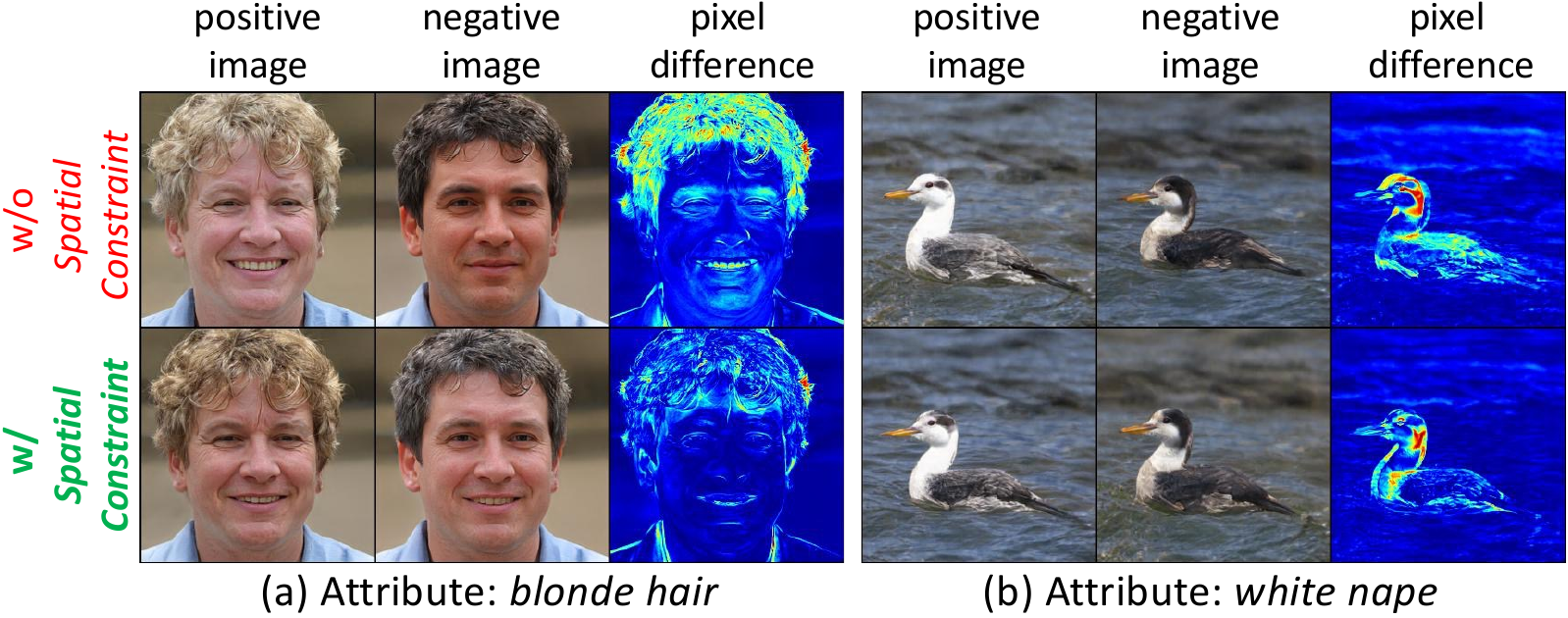}
   \caption{Ablation study of \textit{Spatial Constraint} for identifying attribute directions. Without our \textit{Spatial Constraint} (first row), there are also changes in the other regions (\eg, brows and mouth regions for the \textit{blond hair} attribute; the wings region for the \textit{white nape} attribute). Our \textit{Spatial Constraint} (second row) successfully suppresses the variations in other unintended regions, leading to better disentanglement among different attributes.}
   \label{fig.ablate_spatial_constraint}
   \vspace{-2mm}
\end{figure}

\noindent \textbf{Spatial Constraint} The R-Precision results in \cref{tab.ablate_t2i} show that \textit{Spatial Constraint} can improve the alignment between text and image. The qualitative results in \cref{fig.ablate_spatial_constraint} show that \textit{Spatial Constraint} effectively constrains the spatial variation within the intended region, \eg, hair region for ``\textit{blond hair}'' attribute. These more disentangled attribute directions help StyleT2I achieve better R-Precision performance by adjusting the sentence direction during the inference stage.

\begin{figure}[t]
  \centering
  \includegraphics[width=0.9\linewidth]{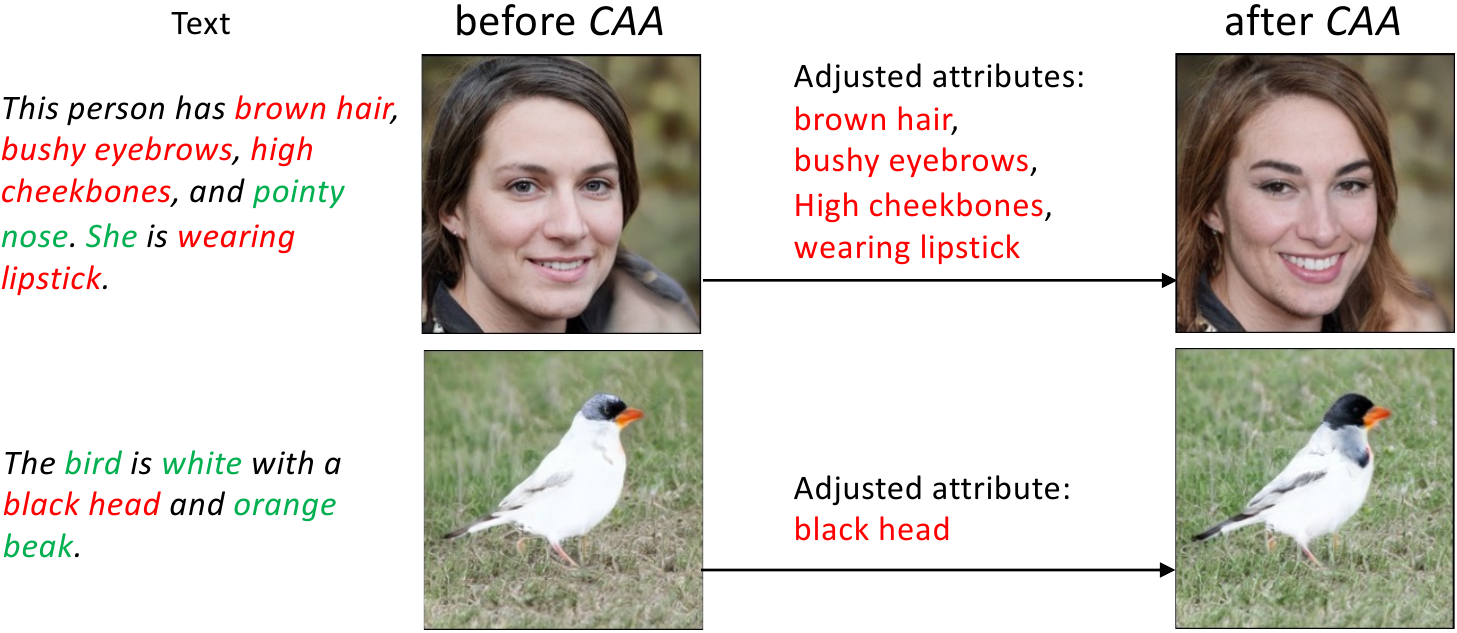}
   \caption{\textit{Compositional Attribute Adjustment} (\textit{CAA}) automatically detect the attributes that are failed to be synthesized (highlighted in \textcolor{red}{red}) and adjust the sentence direction with the attribute directions to improve the compositionality of the text-to-image synthesis results.
   }
   \label{fig.ablate_caa}
   \vspace{-3mm}
\end{figure}

\noindent \textbf{Compositional Attribute Adjustment}
\cref{tab.ablate_t2i} shows that Compositional Attribute Adjustment (\textit{CAA}) improves the R-Precision results and achieves a similar FID result. In \cref{tab.cub_bird_cls}, \textit{CAA} can also improve the unseen bird species classification results, demonstrating its effectiveness for improving compositionality.
In \cref{fig.ablate_caa}, we show that \textit{(CAA)} can not only detects wrong attributes, \eg, ``\textit{brown hair}'', but also correct these wrong attributes by adjusting the sentence direction based on the identified attribute directions.

\noindent \textbf{Finetune CLIP} As introduced in \cref{subsec.implementation_details}, we finetune the CLIP on the training split of the dataset. The R-Precision results in \cref{tab.ablate_t2i} show that finetuning can greatly improve performance. Although trained on a large-scale dataset, the results suggest that CLIP will underperform for text-to-image synthesis with fine-grained attributes, proving the necessity to finetune on the dataset for better results.

\section{Conclusion}
We propose StyleT2I, a new framework for achieving compositional and high-fidelity text-to-image synthesis. We propose a novel \textit{CLIP-guided Contrastive Loss} to better distinguish different compositions, a \textit{Semantic Matching Loss} and a \textit{Spatial Constraint} to identify disentangled attribute directions, and \textit{Compositional Attribute Adjustment} to correct wrong attributes in the synthesis results. StyleT2I outperforms previous approaches in terms of image-text alignment and achieves image fidelity. Admittedly, our work has some limitations. For example, our \textit{Spatial Constraint} is not helpful to disentangle a few attributes that share the same spatial region, \eg, ``\textit{bushy eyebrow}'' and ``\textit{arched eyebrow}.'' One potential negative societal impact is that StyleT2I’s high-fidelity synthesis may be maliciously used for deception. We will mitigate it by asking the users to follow ethical principles when releasing the model. A promising future direction for StyleT2I is complex scene images synthesis for disentangling different objects and backgrounds.

\noindent \textbf{Acknowledgment} \hspace{1pt} This work has been partially supported by the National Science Foundation (NSF) under Grant 1909912 and 1934962 and by the Center of Excellence in Data Science, an Empire State Development-designated Center of Excellence. The article solely reflects the opinions and conclusions of its authors but not the funding agents.

\clearpage
{\small
\bibliographystyle{ieee_fullname}
\bibliography{ref}

\begin{thebibliography}{10}\itemsep=-1pt

\bibitem[Balakrishnan
  \emph{et~al.}(2020)]{balakrishnan2020Eur.Conf.Comput.Vis.ECCV}
Guha Balakrishnan, Yuanjun Xiong, Wei Xia, and Pietro Perona.
\newblock Towards causal benchmarking of bias in face analysis algorithms.
\newblock In {\em The {{European Conference}} on {{Computer Vision}}
  ({{ECCV}})}, 2020.

\bibitem[Bommasani \emph{et~al.}(2021)]{bommasani2021ArXiv210807258Cs}
Rishi Bommasani, Drew~A. Hudson, Ehsan Adeli, Russ Altman, Simran Arora, Sydney
  {von Arx}, Michael~S. Bernstein, Jeannette Bohg, Antoine Bosselut, Emma
  Brunskill, Erik Brynjolfsson, Shyamal Buch, Dallas Card, Rodrigo Castellon,
  Niladri Chatterji, Annie Chen, Kathleen Creel, Jared~Quincy Davis, Dora
  Demszky, Chris Donahue, Moussa Doumbouya, Esin Durmus, Stefano Ermon, John
  Etchemendy, Kawin Ethayarajh, Li {Fei-Fei}, Chelsea Finn, Trevor Gale, Lauren
  Gillespie, Karan Goel, Noah Goodman, Shelby Grossman, Neel Guha, Tatsunori
  Hashimoto, Peter Henderson, John Hewitt, Daniel~E. Ho, Jenny Hong, Kyle Hsu,
  Jing Huang, Thomas Icard, Saahil Jain, Dan Jurafsky, Pratyusha Kalluri,
  Siddharth Karamcheti, Geoff Keeling, Fereshte Khani, Omar Khattab, Pang~Wei
  Koh, Mark Krass, Ranjay Krishna, Rohith Kuditipudi, Ananya Kumar, Faisal
  Ladhak, Mina Lee, Tony Lee, Jure Leskovec, Isabelle Levent, Xiang~Lisa Li,
  Xuechen Li, Tengyu Ma, Ali Malik, Christopher~D. Manning, Suvir Mirchandani,
  Eric Mitchell, Zanele Munyikwa, Suraj Nair, Avanika Narayan, Deepak
  Narayanan, Ben Newman, Allen Nie, Juan~Carlos Niebles, Hamed Nilforoshan,
  Julian Nyarko, Giray Ogut, Laurel Orr, Isabel Papadimitriou, Joon~Sung Park,
  Chris Piech, Eva Portelance, Christopher Potts, Aditi Raghunathan, Rob Reich,
  Hongyu Ren, Frieda Rong, Yusuf Roohani, Camilo Ruiz, Jack Ryan, Christopher
  R{\'e}, Dorsa Sadigh, Shiori Sagawa, Keshav Santhanam, Andy Shih, Krishnan
  Srinivasan, Alex Tamkin, Rohan Taori, Armin~W. Thomas, Florian Tram{\`e}r,
  Rose~E. Wang, William Wang, Bohan Wu, Jiajun Wu, Yuhuai Wu, Sang~Michael Xie,
  Michihiro Yasunaga, Jiaxuan You, Matei Zaharia, Michael Zhang, Tianyi Zhang,
  Xikun Zhang, Yuhui Zhang, Lucia Zheng, Kaitlyn Zhou, and Percy Liang.
\newblock On the {{Opportunities}} and {{Risks}} of {{Foundation Models}}.
\newblock {\em arXiv:2108.07258 [cs]}, 2021.

\bibitem[Buolamwini and
  Gebru(2018)]{buolamwini2018ACMConf.FairnessAccount.Transpar.}
Joy Buolamwini and Timnit Gebru.
\newblock Gender {{Shades}}: {{Intersectional Accuracy Disparities}} in
  {{Commercial Gender Classification}}.
\newblock In {\em {{ACM Conference}} on {{Fairness}}, {{Accountability}}, and
  {{Transparency}}}, 2018.

\bibitem[Chen \emph{et~al.}(2020)]{chen2020Int.Conf.Mach.Learn.b}
Mark Chen, Alec Radford, Rewon Child, Jeff Wu, Heewoo Jun, Prafulla Dhariwal,
  David Luan, and Ilya Sutskever.
\newblock Generative {{Pretraining}} from {{Pixels}}.
\newblock In {\em International {{Conference}} on {{Machine Learning}}}, 2020.

\bibitem[Chen \emph{et~al.}(2018)]{chen2018Adv.NeuralInf.Process.Syst.a}
Ricky T.~Q. Chen, Yulia Rubanova, Jesse Bettencourt, and David~K Duvenaud.
\newblock Neural {{Ordinary Differential Equations}}.
\newblock In {\em Advances in {{Neural Information Processing Systems}}}, 2018.

\bibitem[Chen \emph{et~al.}(2016)]{chen2016Adv.NeuralInf.Process.Syst.}
Xi Chen, Yan Duan, Rein Houthooft, John Schulman, Ilya Sutskever, and Pieter
  Abbeel.
\newblock {{InfoGAN}}: {{Interpretable Representation Learning}} by
  {{Information Maximizing Generative Adversarial Nets}}.
\newblock In {\em Advances in {{Neural Information Processing Systems}}}, 2016.

\bibitem[Chen \emph{et~al.}(2015)]{chen2015ArXiv150400325Cs}
Xinlei Chen, Hao Fang, Tsung-Yi Lin, Ramakrishna Vedantam, Saurabh Gupta, Piotr
  Dollar, and C.~Lawrence Zitnick.
\newblock Microsoft {{COCO Captions}}: {{Data Collection}} and {{Evaluation
  Server}}.
\newblock {\em arXiv:1504.00325 [cs]}, 2015.

\bibitem[Cheng
  \emph{et~al.}(2020)]{cheng2020IEEEConf.Comput.Vis.PatternRecognit.CVPRd}
Bowen Cheng, Maxwell~D. Collins, Yukun Zhu, Ting Liu, Thomas~S. Huang, Hartwig
  Adam, and Liang-Chieh Chen.
\newblock Panoptic-{{DeepLab}}: {{A Simple}}, {{Strong}}, and {{Fast Baseline}}
  for {{Bottom-Up Panoptic Segmentation}}.
\newblock In {\em The {{IEEE Conference}} on {{Computer Vision}} and {{Pattern
  Recognition}} ({{CVPR}})}, 2020.

\bibitem[Chung
  \emph{et~al.}(2014)]{chung2014Adv.NeuralInf.Process.Syst.Workshop}
Junyoung Chung, Caglar Gulcehre, KyungHyun Cho, and Yoshua Bengio.
\newblock Empirical {{Evaluation}} of {{Gated Recurrent Neural Networks}} on
  {{Sequence Modeling}}.
\newblock In {\em Advances in {{Neural Information Processing Systems
  Workshop}}}, 2014.

\bibitem[Dosovitskiy
  \emph{et~al.}(2021)]{dosovitskiy2021Int.Conf.Learn.Represent.}
Alexey Dosovitskiy, Lucas Beyer, Alexander Kolesnikov, Dirk Weissenborn,
  Xiaohua Zhai, Thomas Unterthiner, Mostafa Dehghani, Matthias Minderer, Georg
  Heigold, Sylvain Gelly, Jakob Uszkoreit, and Neil Houlsby.
\newblock An {{Image}} is {{Worth}} 16x16 {{Words}}: {{Transformers}} for
  {{Image Recognition}} at {{Scale}}.
\newblock In {\em International {{Conference}} on {{Learning
  Representations}}}, 2021.

\bibitem[Goodfellow
  \emph{et~al.}(2014)]{goodfellow2014Adv.NeuralInf.Process.Syst.}
Ian Goodfellow, Jean {Pouget-Abadie}, Mehdi Mirza, Bing Xu, David
  {Warde-Farley}, Sherjil Ozair, Aaron Courville, and Yoshua Bengio.
\newblock Generative adversarial nets.
\newblock In {\em Advances in {{Neural Information Processing Systems}}}, 2014.

\bibitem[He
  \emph{et~al.}(2016)]{he2016IEEEConf.Comput.Vis.PatternRecognit.CVPR}
Kaiming He, Xiangyu Zhang, Shaoqing Ren, and Jian Sun.
\newblock Deep residual learning for image recognition.
\newblock In {\em The {{IEEE Conference}} on {{Computer Vision}} and {{Pattern
  Recognition}} ({{CVPR}})}, 2016.

\bibitem[Heusel \emph{et~al.}(2017)]{heusel2017Adv.NeuralInf.Process.Syst.}
Martin Heusel, Hubert Ramsauer, Thomas Unterthiner, Bernhard Nessler, and Sepp
  Hochreiter.
\newblock {{GANs Trained}} by a {{Two Time-Scale Update Rule Converge}} to a
  {{Local Nash Equilibrium}}.
\newblock In {\em Advances in {{Neural Information Processing Systems}}}, 2017.

\bibitem[Higgins \emph{et~al.}(2017)]{higgins2017Int.Conf.Learn.Represent.}
Irina Higgins, Loic Matthey, Arka Pal, Christopher Burgess, Xavier Glorot,
  Matthew Botvinick, Shakir Mohamed, and Alexander Lerchner.
\newblock Beta-{{VAE}}: {{Learning Basic Visual Concepts}} with a {{Constrained
  Variational Framework}}.
\newblock In {\em International {{Conference}} on {{Learning
  Representations}}}, 2017.

\bibitem[Hinz \emph{et~al.}(2020)]{hinz2020IEEETrans.PatternAnal.Mach.Intell.}
Tobias Hinz, Stefan Heinrich, and Stefan Wermter.
\newblock Semantic {{Object Accuracy}} for {{Generative Text-to-Image
  Synthesis}}.
\newblock {\em IEEE Transactions on Pattern Analysis and Machine Intelligence},
  2020.

\bibitem[Honnibal and Montani(2017)]{honnibal2017}
Matthew Honnibal and Ines Montani.
\newblock {{spaCy}} 2: {{Natural}} language understanding with {{Bloom}}
  embeddings, convolutional neural networks and incremental parsing.
\newblock 2017.

\bibitem[Huang
  \emph{et~al.}(2020)]{huang2020IEEEConf.Comput.Vis.PatternRecognit.CVPRb}
Shaofei Huang, Tianrui Hui, Si Liu, Guanbin Li, Yunchao Wei, Jizhong Han, Luoqi
  Liu, and Bo Li.
\newblock Referring {{Image Segmentation}} via {{Cross-Modal Progressive
  Comprehension}}.
\newblock In {\em The {{IEEE Conference}} on {{Computer Vision}} and {{Pattern
  Recognition}} ({{CVPR}})}, 2020.

\bibitem[Karras \emph{et~al.}(2018)]{karras2018Int.Conf.Learn.Represent.}
Tero Karras, Timo Aila, Samuli Laine, and Jaakko Lehtinen.
\newblock Progressive {{Growing}} of {{GANs}} for {{Improved Quality}},
  {{Stability}}, and {{Variation}}.
\newblock In {\em International {{Conference}} on {{Learning
  Representations}}}, 2018.

\bibitem[Karras
  \emph{et~al.}(2019)]{karras2019IEEEConf.Comput.Vis.PatternRecognit.CVPR}
Tero Karras, Samuli Laine, and Timo Aila.
\newblock A {{Style-Based Generator Architecture}} for {{Generative Adversarial
  Networks}}.
\newblock In {\em The {{IEEE Conference}} on {{Computer Vision}} and {{Pattern
  Recognition}} ({{CVPR}})}, 2019.

\bibitem[Karras
  \emph{et~al.}(2020)]{karras2020IEEEConf.Comput.Vis.PatternRecognit.CVPR}
Tero Karras, Samuli Laine, Miika Aittala, Janne Hellsten, Jaakko Lehtinen, and
  Timo Aila.
\newblock Analyzing and {{Improving}} the {{Image Quality}} of {{StyleGAN}}.
\newblock In {\em The {{IEEE Conference}} on {{Computer Vision}} and {{Pattern
  Recognition}} ({{CVPR}})}, 2020.

\bibitem[Kim \emph{et~al.}(2018)]{kim2018Int.Conf.Mach.Learn.a}
Been Kim, Martin Wattenberg, Justin Gilmer, Carrie Cai, James Wexler, Fernanda
  Viegas, and Rory Sayres.
\newblock Interpretability {{Beyond Feature Attribution}}: {{Quantitative
  Testing}} with {{Concept Activation Vectors}} ({{TCAV}}).
\newblock In {\em International {{Conference}} on {{Machine Learning}}}, 2018.

\bibitem[Kingma and Ba(2015)]{kingma2015Int.Conf.Learn.Represent.}
Diederik~P. Kingma and Jimmy Ba.
\newblock Adam: {{A Method}} for {{Stochastic Optimization}}.
\newblock In {\em International {{Conference}} on {{Learning
  Representations}}}, 2015.

\bibitem[Kingma and Dhariwal(2018)]{kingma2018Adv.NeuralInf.Process.Syst.}
Durk~P Kingma and Prafulla Dhariwal.
\newblock Glow: {{Generative Flow}} with {{Invertible}} 1x1 {{Convolutions}}.
\newblock In {\em Advances in {{Neural Information Processing Systems}}}, 2018.

\bibitem[Kingma and Welling(2014)]{kingma2014Int.Conf.Learn.Represent.}
Diederik~P. Kingma and Max Welling.
\newblock Auto-encoding variational bayes.
\newblock In {\em International {{Conference}} on {{Learning
  Representations}}}, 2014.

\bibitem[Koh \emph{et~al.}(2021)]{koh2021IEEEWinterConf.Appl.Comput.Vis.WACV}
Jing~Yu Koh, Jason Baldridge, Honglak Lee, and Yinfei Yang.
\newblock Text-to-{{Image Generation Grounded}} by {{Fine-Grained User
  Attention}}.
\newblock In {\em The {{IEEE Winter Conference}} on {{Applications}} of
  {{Computer Vision}} ({{WACV}})}, 2021.

\bibitem[Kumar \emph{et~al.}(2018)]{kumar2018Int.Conf.Learn.Represent.}
Abhishek Kumar, Prasanna Sattigeri, and Avinash Balakrishnan.
\newblock Variational {{Inference}} of {{Disentangled Latent Concepts}} from
  {{Unlabeled Observations}}.
\newblock In {\em International {{Conference}} on {{Learning
  Representations}}}, 2018.

\bibitem[Lang \emph{et~al.}(2021)]{lang2021IEEEInt.Conf.Comput.Vis.ICCV}
Oran Lang, Yossi Gandelsman, Michal Yarom, Yoav Wald, Gal Elidan, Avinatan
  Hassidim, William~T. Freeman, Phillip Isola, Amir Globerson, Michal Irani,
  and Inbar Mosseri.
\newblock Explaining in {{Style}}: {{Training}} a {{GAN}} to explain a
  classifier in {{StyleSpace}}.
\newblock In {\em The {{IEEE International Conference}} on {{Computer Vision}}
  ({{ICCV}})}, 2021.

\bibitem[Lee
  \emph{et~al.}(2020)]{lee2020IEEEConf.Comput.Vis.PatternRecognit.CVPR}
Hyodong Lee, Joonseok Lee, Joe Yue-Hei Ng, and Paul Natsev.
\newblock Large {{Scale Video Representation Learning}} via {{Relational Graph
  Clustering}}.
\newblock In {\em The {{IEEE Conference}} on {{Computer Vision}} and {{Pattern
  Recognition}} ({{CVPR}})}, 2020.

\bibitem[Li
  \emph{et~al.}(2019)]{li2019IEEEConf.Comput.Vis.PatternRecognit.CVPRf}
Aoxue Li, Tiange Luo, Zhiwu Lu, Tao Xiang, and Liwei Wang.
\newblock Large-{{Scale Few-Shot Learning}}: {{Knowledge Transfer With Class
  Hierarchy}}.
\newblock In {\em The {{IEEE Conference}} on {{Computer Vision}} and {{Pattern
  Recognition}} ({{CVPR}})}, 2019.

\bibitem[Li \emph{et~al.}(2019)]{li2019Adv.NeuralInf.Process.Syst.b}
Bowen Li, Xiaojuan Qi, Thomas Lukasiewicz, and Philip Torr.
\newblock Controllable {{Text-to-Image Generation}}.
\newblock In {\em Advances in {{Neural Information Processing Systems}}}, 2019.

\bibitem[Li and Xu(2021)]{li2021IEEEInt.Conf.Comput.Vis.ICCV}
Zhiheng Li and Chenliang Xu.
\newblock Discover the {{Unknown Biased Attribute}} of an {{Image Classifier}}.
\newblock In {\em The {{IEEE International Conference}} on {{Computer Vision}}
  ({{ICCV}})}, 2021.

\bibitem[Liang \emph{et~al.}(2020)]{liang2020Eur.Conf.Comput.Vis.ECCV}
Jiadong Liang, Wenjie Pei, and Feng Lu.
\newblock {{CPGAN}}: {{Content-Parsing Generative Adversarial Networks}} for
  {{Text-to-Image Synthesis}}.
\newblock In {\em The {{European Conference}} on {{Computer Vision}}
  ({{ECCV}})}, 2020.

\bibitem[Lin \emph{et~al.}(2014)]{lin2014Eur.Conf.Comput.Vis.ECCV}
Tsung-Yi Lin, Michael Maire, Serge Belongie, James Hays, Pietro Perona, Deva
  Ramanan, Piotr Doll{\'a}r, and C.~Lawrence Zitnick.
\newblock Microsoft {{COCO}}: {{Common Objects}} in {{Context}}.
\newblock In {\em The {{European Conference}} on {{Computer Vision}}
  ({{ECCV}})}, 2014.

\bibitem[Liu \emph{et~al.}(2015)]{liu2015IEEEInt.Conf.Comput.Vis.ICCV}
Ziwei Liu, Ping Luo, Xiaogang Wang, and Xiaoou Tang.
\newblock Deep {{Learning Face Attributes}} in the {{Wild}}.
\newblock In {\em The {{IEEE International Conference}} on {{Computer Vision}}
  ({{ICCV}})}, 2015.

\bibitem[Locatello \emph{et~al.}(2019)]{locatello2019Int.Conf.Mach.Learn.}
Francesco Locatello, Stefan Bauer, Mario Lucic, Gunnar Raetsch, Sylvain Gelly,
  Bernhard Sch{\"o}lkopf, and Olivier Bachem.
\newblock Challenging {{Common Assumptions}} in the {{Unsupervised Learning}}
  of {{Disentangled Representations}}.
\newblock In {\em International {{Conference}} on {{Machine Learning}}}, 2019.

\bibitem[Loshchilov and Hutter(2019)]{loshchilov2019Int.Conf.Learn.Represent.}
Ilya Loshchilov and Frank Hutter.
\newblock Decoupled {{Weight Decay Regularization}}.
\newblock In {\em International {{Conference}} on {{Learning
  Representations}}}, 2019.

\bibitem[Nair and Hinton(2010)]{nair2010Int.Conf.Mach.Learn.}
Vinod Nair and Geoffrey~E. Hinton.
\newblock Rectified linear units improve restricted boltzmann machines.
\newblock In {\em International {{Conference}} on {{Machine Learning}}}, 2010.

\bibitem[Otsu(1979)]{otsu1979IEEETrans.Syst.ManCybern.}
Nobuyuki Otsu.
\newblock A {{Threshold Selection Method}} from {{Gray-Level Histograms}}.
\newblock {\em IEEE Transactions on Systems, Man, and Cybernetics}, 1979.

\bibitem[Park
  \emph{et~al.}(2021)]{park2021Thirty-FifthConf.NeuralInf.Process.Syst.DatasetsBenchmarksTrackRound1}
Dong~Huk Park, Samaneh Azadi, Xihui Liu, Trevor Darrell, and Anna Rohrbach.
\newblock Benchmark for {{Compositional Text-to-Image Synthesis}}.
\newblock In {\em Thirty-Fifth {{Conference}} on {{Neural Information
  Processing Systems Datasets}} and {{Benchmarks Track}} ({{Round}} 1)}, 2021.

\bibitem[Park
  \emph{et~al.}(2019)]{park2019IEEEConf.Comput.Vis.PatternRecognit.CVPR}
Taesung Park, Ming-Yu Liu, Ting-Chun Wang, and Jun-Yan Zhu.
\newblock Semantic {{Image Synthesis With Spatially-Adaptive Normalization}}.
\newblock In {\em The {{IEEE Conference}} on {{Computer Vision}} and {{Pattern
  Recognition}} ({{CVPR}})}, 2019.

\bibitem[Paszke \emph{et~al.}(2019)]{paszke2019Adv.NeuralInf.Process.Syst.}
Adam Paszke, Sam Gross, Francisco Massa, Adam Lerer, James Bradbury, Gregory
  Chanan, Trevor Killeen, Zeming Lin, Natalia Gimelshein, Luca Antiga, Alban
  Desmaison, Andreas Kopf, Edward Yang, Zachary DeVito, Martin Raison, Alykhan
  Tejani, Sasank Chilamkurthy, Benoit Steiner, Lu Fang, Junjie Bai, and Soumith
  Chintala.
\newblock {{PyTorch}}: {{An Imperative Style}}, {{High-Performance Deep
  Learning Library}}.
\newblock In {\em Advances in {{Neural Information Processing Systems}}}, 2019.

\bibitem[Patashnik
  \emph{et~al.}(2021)]{patashnik2021IEEEInt.Conf.Comput.Vis.ICCV}
Or Patashnik, Zongze Wu, Eli Shechtman, Daniel {Cohen-Or}, and Dani Lischinski.
\newblock {{StyleCLIP}}: {{Text-Driven Manipulation}} of {{StyleGAN Imagery}}.
\newblock In {\em The {{IEEE International Conference}} on {{Computer Vision}}
  ({{ICCV}})}, 2021.

\bibitem[Peebles \emph{et~al.}(2020)]{peebles2020Eur.Conf.Comput.Vis.ECCV}
William Peebles, John Peebles, Jun-Yan Zhu, Alexei Efros, and Antonio Torralba.
\newblock The {{Hessian Penalty}}: {{A Weak Prior}} for {{Unsupervised
  Disentanglement}}.
\newblock In {\em The {{European Conference}} on {{Computer Vision}}
  ({{ECCV}})}, 2020.

\bibitem[Pennington
  \emph{et~al.}(2014)]{pennington2014Empir.MethodsNat.Lang.Process.}
Jeffrey Pennington, Richard Socher, and Christopher Manning.
\newblock {{GloVe}}: {{Global Vectors}} for {{Word Representation}}.
\newblock In {\em Empirical {{Methods}} in {{Natural Language Processing}}},
  2014.

\bibitem[Prabhu \emph{et~al.}(2019)]{prabhu2019Int.Conf.Mach.Learn.Workshop}
Vinay~Uday Prabhu, Dian~Ang Yap, Alexander Wang, and John Whaley.
\newblock Covering up bias in {{CelebA-like}} datasets with {{Markov}}
  blankets: {{A}} post-hoc cure for attribute prior avoidance.
\newblock In {\em International {{Conference}} on {{Machine Learning
  Workshop}}}, 2019.

\bibitem[Qiao
  \emph{et~al.}(2019)]{qiao2019IEEEConf.Comput.Vis.PatternRecognit.CVPRa}
Tingting Qiao, Jing Zhang, Duanqing Xu, and Dacheng Tao.
\newblock {{MirrorGAN}}: {{Learning Text-To-Image Generation}} by
  {{Redescription}}.
\newblock In {\em The {{IEEE Conference}} on {{Computer Vision}} and {{Pattern
  Recognition}} ({{CVPR}})}, 2019.

\bibitem[Radford \emph{et~al.}(2021)]{radford2021Int.Conf.Mach.Learn.}
Alec Radford, Jong~Wook Kim, Chris Hallacy, Aditya Ramesh, Gabriel Goh,
  Sandhini Agarwal, Girish Sastry, Amanda Askell, Pamela Mishkin, Jack Clark,
  Gretchen Krueger, and Ilya Sutskever.
\newblock Learning {{Transferable Visual Models From Natural Language
  Supervision}}.
\newblock In {\em International {{Conference}} on {{Machine Learning}}}, 2021.

\bibitem[Ramesh \emph{et~al.}(2021)]{ramesh2021Int.Conf.Mach.Learn.}
Aditya Ramesh, Mikhail Pavlov, Gabriel Goh, Scott Gray, Chelsea Voss, Alec
  Radford, Mark Chen, and Ilya Sutskever.
\newblock Zero-{{Shot Text-to-Image Generation}}.
\newblock In {\em International {{Conference}} on {{Machine Learning}}}, 2021.

\bibitem[Reed
  \emph{et~al.}(2016)]{reed2016IEEEConf.Comput.Vis.PatternRecognit.CVPR}
Scott Reed, Zeynep Akata, Honglak Lee, and Bernt Schiele.
\newblock Learning {{Deep Representations}} of {{Fine-Grained Visual
  Descriptions}}.
\newblock In {\em The {{IEEE Conference}} on {{Computer Vision}} and {{Pattern
  Recognition}} ({{CVPR}})}, 2016.

\bibitem[Reed \emph{et~al.}(2016)]{reed2016Int.Conf.Mach.Learn.}
Scott Reed, Zeynep Akata, Xinchen Yan, Lajanugen Logeswaran, Bernt Schiele, and
  Honglak Lee.
\newblock Generative {{Adversarial Text}} to {{Image Synthesis}}.
\newblock In {\em International {{Conference}} on {{Machine Learning}}}, 2016.

\bibitem[Ruan \emph{et~al.}(2021)]{ruan2021IEEEInt.Conf.Comput.Vis.ICCV}
Shulan Ruan, Yong Zhang, Kun Zhang, Yanbo Fan, Fan Tang, Qi Liu, and Enhong
  Chen.
\newblock {{DAE-GAN}}: {{Dynamic Aspect-aware GAN}} for {{Text-to-Image
  Synthesis}}.
\newblock In {\em The {{IEEE International Conference}} on {{Computer Vision}}
  ({{ICCV}})}, 2021.

\bibitem[Schroff
  \emph{et~al.}(2015)]{schroff2015IEEEConf.Comput.Vis.PatternRecognit.CVPR}
Florian Schroff, Dmitry Kalenichenko, and James Philbin.
\newblock {{FaceNet}}: {{A Unified Embedding}} for {{Face Recognition}} and
  {{Clustering}}.
\newblock In {\em The {{IEEE Conference}} on {{Computer Vision}} and {{Pattern
  Recognition}} ({{CVPR}})}, 2015.

\bibitem[Shen
  \emph{et~al.}(2020)]{shen2020IEEEConf.Comput.Vis.PatternRecognit.CVPR}
Yujun Shen, Jinjin Gu, Xiaoou Tang, and Bolei Zhou.
\newblock Interpreting the {{Latent Space}} of {{GANs}} for {{Semantic Face
  Editing}}.
\newblock In {\em The {{IEEE Conference}} on {{Computer Vision}} and {{Pattern
  Recognition}} ({{CVPR}})}, 2020.

\bibitem[Shen \emph{et~al.}(2020)]{shen2020IEEETrans.PatternAnal.Mach.Intell.}
Yujun Shen, Ceyuan Yang, Xiaoou Tang, and Bolei Zhou.
\newblock {{InterFaceGAN}}: {{Interpreting}} the {{Disentangled Face
  Representation Learned}} by {{GANs}}.
\newblock {\em IEEE Transactions on Pattern Analysis and Machine Intelligence},
  2020.

\bibitem[Shi
  \emph{et~al.}(2022)]{shi2022IEEEConf.Comput.Vis.PatternRecognit.CVPR}
Jing Shi, Ning Xu, Haitian Zheng, Alex Smith, Jiebo Luo, and Chenliang Xu.
\newblock {{SpaceEdit}}: {{Learning}} a {{Unified Editing Space}} for
  {{Open-domain Image Color Editing}}.
\newblock In {\em The {{IEEE Conference}} on {{Computer Vision}} and {{Pattern
  Recognition}} ({{CVPR}})}, 2022.

\bibitem[Sohn(2016)]{sohn2016Adv.NeuralInf.Process.Syst.}
Kihyuk Sohn.
\newblock Improved {{Deep Metric Learning}} with {{Multi-class N-pair Loss
  Objective}}.
\newblock In {\em Advances in {{Neural Information Processing Systems}}}, 2016.

\bibitem[Tan
  \emph{et~al.}(2019)]{tan2019IEEEConf.Comput.Vis.PatternRecognit.CVPR}
Fuwen Tan, Song Feng, and Vicente Ordonez.
\newblock {{Text2Scene}}: {{Generating Compositional Scenes From Textual
  Descriptions}}.
\newblock In {\em The {{IEEE Conference}} on {{Computer Vision}} and {{Pattern
  Recognition}} ({{CVPR}})}, 2019.

\bibitem[{van den Oord}
  \emph{et~al.}(2017)]{vandenoord2017Adv.NeuralInf.Process.Syst.}
Aaron {van den Oord}, Oriol Vinyals, and koray {kavukcuoglu}.
\newblock Neural {{Discrete Representation Learning}}.
\newblock In {\em Advances in {{Neural Information Processing Systems}}}, 2017.

\bibitem[Van~Horn
  \emph{et~al.}(2015)]{vanhorn2015IEEEConf.Comput.Vis.PatternRecognit.CVPR}
Grant Van~Horn, Steve Branson, Ryan Farrell, Scott Haber, Jessie Barry, Panos
  Ipeirotis, Pietro Perona, and Serge Belongie.
\newblock Building a {{Bird Recognition App}} and {{Large Scale Dataset With
  Citizen Scientists}}: {{The Fine Print}} in {{Fine-Grained Dataset
  Collection}}.
\newblock In {\em The {{IEEE Conference}} on {{Computer Vision}} and {{Pattern
  Recognition}} ({{CVPR}})}, 2015.

\bibitem[Vaswani \emph{et~al.}(2017)]{vaswani2017Adv.NeuralInf.Process.Syst.}
Ashish Vaswani, Noam Shazeer, Niki Parmar, Jakob Uszkoreit, Llion Jones,
  Aidan~N. Gomez, {\L}ukasz Kaiser, and Illia Polosukhin.
\newblock Attention is {{All}} you {{Need}}.
\newblock In {\em Advances in {{Neural Information Processing Systems}}}, 2017.

\bibitem[Wah \emph{et~al.}(2011)]{wah2011}
C. Wah, S. Branson, P. Welinder, P. Perona, and S. Belongie.
\newblock The {{Caltech-UCSD Birds-200-2011 Dataset}}.
\newblock Technical report, {California Institute of Technology}, 2011.

\bibitem[Wang
  \emph{et~al.}(2021)]{wang2021IEEEWinterConf.Appl.Comput.Vis.WACVa}
Tianren Wang, Teng Zhang, and Brian Lovell.
\newblock Faces a la {{Carte}}: {{Text-to-Face Generation}} via {{Attribute
  Disentanglement}}.
\newblock In {\em The {{IEEE Winter Conference}} on {{Applications}} of
  {{Computer Vision}} ({{WACV}})}, 2021.

\bibitem[Wei \emph{et~al.}(2021)]{wei2021IEEEInt.Conf.Comput.Vis.ICCV}
Yuxiang Wei, Yupeng Shi, Xiao Liu, Zhilong Ji, Yuan Gao, Zhongqin Wu, and
  Wangmeng Zuo.
\newblock Orthogonal {{Jacobian Regularization}} for {{Unsupervised
  Disentanglement}} in {{Image Generation}}.
\newblock In {\em The {{IEEE International Conference}} on {{Computer Vision}}
  ({{ICCV}})}, 2021.

\bibitem[Xia
  \emph{et~al.}(2021)]{xia2021IEEEConf.Comput.Vis.PatternRecognit.CVPR}
Weihao Xia, Yujiu Yang, Jing-Hao Xue, and Baoyuan Wu.
\newblock {{TediGAN}}: {{Text-Guided Diverse Face Image Generation}} and
  {{Manipulation}}.
\newblock In {\em The {{IEEE Conference}} on {{Computer Vision}} and {{Pattern
  Recognition}} ({{CVPR}})}, 2021.

\bibitem[Xia \emph{et~al.}(2021)]{xia2021ArXiv210408910Cs}
Weihao Xia, Yujiu Yang, Jing-Hao Xue, and Baoyuan Wu.
\newblock Towards {{Open-World Text-Guided Face Image Generation}} and
  {{Manipulation}}.
\newblock {\em arXiv:2104.08910 [cs]}, 2021.

\bibitem[Xian \emph{et~al.}(2019)]{xian2019IEEETrans.PatternAnal.Mach.Intell.}
Yongqin Xian, Christoph~H. Lampert, Bernt Schiele, and Zeynep Akata.
\newblock Zero-{{Shot Learning}}\textemdash{{A Comprehensive Evaluation}} of
  the {{Good}}, the {{Bad}} and the {{Ugly}}.
\newblock {\em IEEE Transactions on Pattern Analysis and Machine Intelligence},
  2019.

\bibitem[Xian
  \emph{et~al.}(2017)]{xian2017IEEEConf.Comput.Vis.PatternRecognit.CVPR}
Yongqin Xian, Bernt Schiele, and Zeynep Akata.
\newblock Zero-{{Shot Learning}} \textemdash{} {{The Good}}, the {{Bad}} and
  the {{Ugly}}.
\newblock In {\em The {{IEEE Conference}} on {{Computer Vision}} and {{Pattern
  Recognition}} ({{CVPR}})}, 2017.

\bibitem[Xu
  \emph{et~al.}(2018)]{xu2018IEEEConf.Comput.Vis.PatternRecognit.CVPRa}
Tao Xu, Pengchuan Zhang, Qiuyuan Huang, Han Zhang, Zhe Gan, Xiaolei Huang, and
  Xiaodong He.
\newblock {{AttnGAN}}: {{Fine-Grained Text}} to {{Image Generation With
  Attentional Generative Adversarial Networks}}.
\newblock In {\em The {{IEEE Conference}} on {{Computer Vision}} and {{Pattern
  Recognition}} ({{CVPR}})}, 2018.

\bibitem[Yao \emph{et~al.}(2021)]{yao2021IEEEInt.Conf.Comput.Vis.ICCV}
Xu Yao, Alasdair Newson, Yann Gousseau, and Pierre Hellier.
\newblock A {{Latent Transformer}} for {{Disentangled Face Editing}} in
  {{Images}} and {{Videos}}.
\newblock In {\em The {{IEEE International Conference}} on {{Computer Vision}}
  ({{ICCV}})}, 2021.

\bibitem[Yin
  \emph{et~al.}(2019)]{yin2019IEEEConf.Comput.Vis.PatternRecognit.CVPR}
Guojun Yin, Bin Liu, Lu Sheng, Nenghai Yu, Xiaogang Wang, and Jing Shao.
\newblock Semantics {{Disentangling}} for {{Text-to-Image Generation}}.
\newblock In {\em The {{IEEE Conference}} on {{Computer Vision}} and {{Pattern
  Recognition}} ({{CVPR}})}, 2019.

\bibitem[Zhang
  \emph{et~al.}(2021)]{zhang2021IEEEConf.Comput.Vis.PatternRecognit.CVPRa}
Han Zhang, Jing~Yu Koh, Jason Baldridge, Honglak Lee, and Yinfei Yang.
\newblock Cross-{{Modal Contrastive Learning}} for {{Text-to-Image
  Generation}}.
\newblock In {\em The {{IEEE Conference}} on {{Computer Vision}} and {{Pattern
  Recognition}} ({{CVPR}})}, 2021.

\bibitem[Zhang \emph{et~al.}(2017)]{zhang2017IEEEInt.Conf.Comput.Vis.ICCVa}
Han Zhang, Tao Xu, Hongsheng Li, Shaoting Zhang, Xiaogang Wang, Xiaolei Huang,
  and Dimitris~N. Metaxas.
\newblock {{StackGAN}}: {{Text}} to {{Photo-Realistic Image Synthesis With
  Stacked Generative Adversarial Networks}}.
\newblock In {\em The {{IEEE International Conference}} on {{Computer Vision}}
  ({{ICCV}})}, 2017.

\bibitem[Zhang
  \emph{et~al.}(2019)]{zhang2019IEEETrans.PatternAnal.Mach.Intell.}
Han Zhang, Tao Xu, Hongsheng Li, Shaoting Zhang, Xiaogang Wang, Xiaolei Huang,
  and Dimitris~N. Metaxas.
\newblock {{StackGAN}}++: {{Realistic Image Synthesis}} with {{Stacked
  Generative Adversarial Networks}}.
\newblock {\em IEEE Transactions on Pattern Analysis and Machine Intelligence},
  2019.

\bibitem[Zhang
  \emph{et~al.}(2021)]{zhang2021IEEEConf.Comput.Vis.PatternRecognit.CVPR}
Yuxuan Zhang, Huan Ling, Jun Gao, Kangxue Yin, Jean-Francois Lafleche, Adela
  Barriuso, Antonio Torralba, and Sanja Fidler.
\newblock {{DatasetGAN}}: {{Efficient Labeled Data Factory}} with {{Minimal
  Human Effort}}.
\newblock In {\em The {{IEEE Conference}} on {{Computer Vision}} and {{Pattern
  Recognition}} ({{CVPR}})}, 2021.

\bibitem[Zhu
  \emph{et~al.}(2019)]{zhu2019IEEEConf.Comput.Vis.PatternRecognit.CVPRc}
Minfeng Zhu, Pingbo Pan, Wei Chen, and Yi Yang.
\newblock {{DM-GAN}}: {{Dynamic Memory Generative Adversarial Networks}} for
  {{Text-To-Image Synthesis}}.
\newblock In {\em The {{IEEE Conference}} on {{Computer Vision}} and {{Pattern
  Recognition}} ({{CVPR}})}, 2019.

\bibitem[Zhu
  \emph{et~al.}(2021)]{zhu2021IEEEConf.Comput.Vis.PatternRecognit.CVPR}
Xinqi Zhu, Chang Xu, and Dacheng Tao.
\newblock Where and {{What}}? {{Examining Interpretable Disentangled
  Representations}}.
\newblock In {\em The {{IEEE Conference}} on {{Computer Vision}} and {{Pattern
  Recognition}} ({{CVPR}})}, 2021.

\bibitem[Zhuang \emph{et~al.}(2021)]{zhuang2021Int.Conf.Learn.Represent.}
Peiye Zhuang, Oluwasanmi Koyejo, and Alexander~G. Schwing.
\newblock Enjoy {{Your Editing}}: {{Controllable GANs}} for {{Image Editing}}
  via {{Latent Space Navigation}}.
\newblock In {\em International {{Conference}} on {{Learning
  Representations}}}, 2021.

\end{thebibliography}
}

\clearpage

\section*{Appendix}

\appendix

\section{Implementation Details}

\subsection{Complete Algorithm}
\label{subsec.supp.complete_algo}

Training the StyleT2I framework contains two steps---Step 1: train the \textit{Text-to-Direction} module (\cref{alg.supp.train_t2d}); Step 2: train the \textit{Attribute-to-Direction} module (\cref{alg.supp.train_a2d}).
The pseudocode of the inference algorithm of StyleT2I for synthesizing images conditioned on the given text is shown in \cref{alg.supp.inference_stylet2i}.

\begin{algorithm}
  \caption{Train \textit{Text-to-Direction} module}
  \label{alg.supp.train_t2d}
  \DontPrintSemicolon
    \KwInput{$G$: pretrained generator, $M_t$: training iterations, $\mathcal{T} = \{ \mathbf{t} \}$: training set of text.}
    \KwOutput{$\mathcal{F}_\text{text}$: \textit{Text-to-Direction} module}
    \For{$k : 1 \dots M_t$}
    {
        $\mathbf{z} \sim \mathcal{W}+$ \tcp*{random latent code sampled from $\mathcal{W}+$ space}
        $\mathbf{t} \sim \mathcal{T}$ \tcp*{text sampled from the training set}
        $\mathbf{s} = \mathcal{F}_\text{text}(\mathbf{z}, \mathbf{t})$ \tcp*{predict sentence direction}
        $\mathbf{z}_\mathbf{s} = \mathbf{z} + \mathbf{s}$ \tcp*{text-conditioned code}
        $\mathbf{\hat{I}} = G(\mathbf{z}_\mathbf{s})$ \tcp*{synthesize image}
        $\mathcal{L}_s = \mathcal{L}_\text{contras}(\mathbf{\hat{I}}, \mathbf{t}) + \mathcal{L}_\text{norm}(\mathbf{s}) $ \tcp*{compute loss}
        $\mathcal{F}_\text{text} \leftarrow \text{Adam}(\nabla_{\mathcal{F}_\text{text}} \mathcal{L}_s)$ \tcp*{update $\mathcal{F}_\text{text}$}
    }
     \textbf{return} $\mathcal{F}_\text{text}$
\end{algorithm}

\begin{algorithm}
  \caption{Train \textit{Attribute-to-Direction} module}
  \label{alg.supp.train_a2d}
  \DontPrintSemicolon
    \KwInput{$\mathcal{V} = \{ \mathbf{t}^a \}$: attribute vocabulary, $G$: pretrained generator, $\mathcal{S}$: weakly-supervised segmentation network, $M_a$: training iterations}
    \KwOutput{$\mathcal{F}_\text{attr}$: \textit{Attribute-to-Direction} module}
    \For{$m : 1 \dots M_a$}
    {
        $\mathbf{z} \sim \mathcal{W}+$ \tcp*{random latent code sampled from $\mathcal{W}+$ space}
        $\mathbf{t}^a \sim \mathcal{V}$ \tcp*{attribute sampled from vocabulary}
        $\mathbf{a} = \mathcal{F}_\text{attr}(\mathbf{z}, \mathbf{t}^a)$ \tcp*{predict attribute direction}
        $\mathbf{z}_\text{pos}^a = \mathbf{z} + \mathbf{a}$ \tcp*{positive latent code}
        $\mathbf{z}_\text{neg}^a = \mathbf{z} - \mathbf{a}$ \tcp*{negative latent code}
        $\mathbf{I}_\text{pos}^a = G(\mathbf{z}_\text{pos}^a)$ \tcp*{positive image}
        $\mathbf{I}_\text{neg}^a = G(\mathbf{z}_\text{neg}^a)$ \tcp*{negative image}
        $\mathbf{M}^a = \mathcal{S}(\mathbf{I}_\text{pos}^a)$ \tcp*{pseudo-ground-truth mask}
        $\mathbf{I}_\text{diff}^a = \sum_c |\mathbf{I}_\text{pos}^a - \mathbf{I}_\text{neg}^a|$ \tcp*{pixel-level difference}
        $\mathbf{\tilde{I}}_\text{diff}^a = \frac{\mathbf{I}_\text{diff}^a - \text{min}(\mathbf{I}_\text{diff}^a)}{\text{max}(\mathbf{I}_\text{diff}^a) - \text{min}(\mathbf{I}_\text{diff}^a)}$ \tcp*{min-max normalization}
        $\mathcal{L}_a = \mathcal{L}_\text{semantic}(\mathbf{I}_\text{pos}^a, \mathbf{I}_\text{neg}^a, \mathbf{t}^a) + \mathcal{L}_\text{spatial}(\mathbf{\tilde{I}}_\text{diff}^a, \mathbf{M}^a) + \mathcal{L}_\text{norm}(\mathbf{a}) $ \tcp*{compute loss}
        $\mathcal{F}_\text{attr} \leftarrow \text{Adam}(\nabla_{\mathcal{F}_\text{attr}} \mathcal{L}_a)$ \tcp*{update $\mathcal{F}_\text{attr}$}
    }
     \textbf{return} $\mathcal{F}_\text{attr}$
\end{algorithm}

\begin{algorithm}
  \caption{Inference algorithm of StyleT2I}
  \label{alg.supp.inference_stylet2i}
  \DontPrintSemicolon
    \KwInput{$G$: pretrained generator, $\mathbf{t}$: input text, $\{\mathbf{t}_i^a\}_{i=1}^K$: extracted $K$ attributes from text, $\mathcal{F}_\text{text}$: \textit{Text-to-Direction} module, $\mathcal{F}_\text{attr}$: \textit{Attribute-to-Direction} module}
    \KwOutput{$\mathbf{\hat{I}}$: synthesized image conditioned on the input text}
    $\mathbf{z} \sim \mathcal{W}+$ \tcp*{random latent code sampled from $\mathcal{W}+$ space}
    $\mathbf{s} = \mathcal{F}_\text{text}(\mathbf{z}, \mathbf{t})$ \tcp*{predict sentence direction}
    $\mathbf{A} = \{ \mathbf{a}_i \mid \cos(\mathbf{a}_i, \mathbf{s}) \leq 0 \}$. \tcp*{set of attributes need to be adjusted}
    $\mathbf{s}' = \mathbf{s} + \sum_{\mathbf{a}_i \in \mathbf{A}} \frac{\mathbf{a}_i}{||\mathbf{a}_i||_2}$ \tcp*{adjust sentence direction}
    $\mathbf{z}_\mathbf{s} = \mathbf{z} + \mathbf{s}'$ \tcp*{text-conditioned code}
    $\mathbf{\hat{I}} = G(\mathbf{z}_\mathbf{s})$ \tcp*{synthesize image}
    \textbf{return} $\mathbf{\hat{I}}$
\end{algorithm}

\subsection{Hyperparameters and Network Architecture}
\label{subsec.supp.hyperparameters_network_arch}

\begin{figure}[b]
    \centering
    \includegraphics[width=\linewidth]{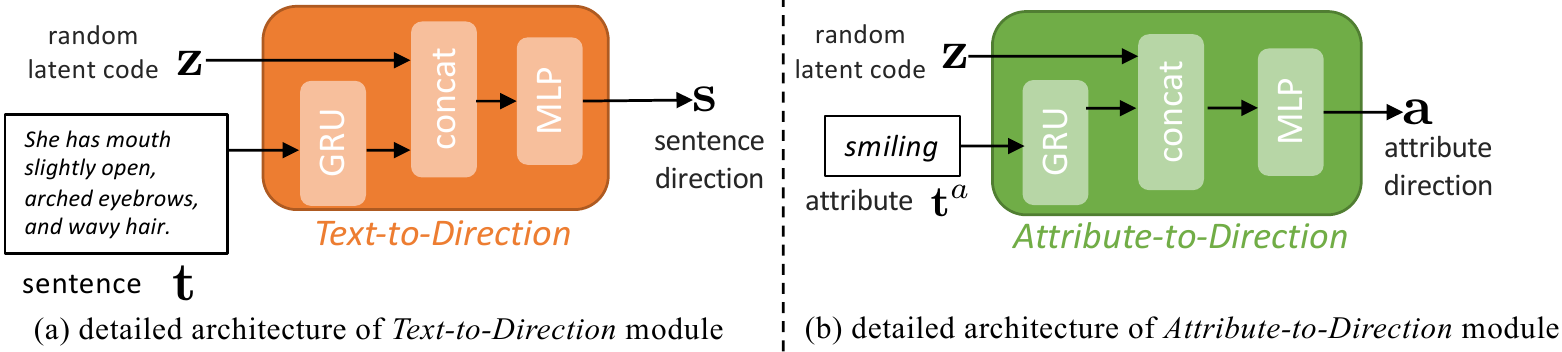}
    \caption{Detailed architectures of (a) \textit{Text-to-Direction} module and (b) \textit{Attribute-to-Direction} module.}
    \label{fig.supp.arch}
\end{figure}

We pretrain StyleGAN2 on each dataset (CelebA-HQ~\cite{karras2018Int.Conf.Learn.Represent.} and CUB~\cite{wah2011}) with 300,000 iterations. In CLIP~\cite{radford2021Int.Conf.Mach.Learn.}, we use ViT-B/32~\cite{dosovitskiy2021Int.Conf.Learn.Represent.} architecture as the image encoder. We use Adam optimizer~\cite{kingma2015Int.Conf.Learn.Represent.} with $10^{-4}$ learning rate to train both modules. The \textit{Text-to-Direction} module is trained with 60,000 iterations and the batch size is 40. The \textit{Attribute-to-Direction} module is trained with 1000 iterations with batch size of 2. The architectures of \textit{Text-to-Direction} module and \textit{Attribute-to-Direction} module are shown in \cref{fig.supp.arch}.

\subsection{Attribute Extraction}
\label{subsec.attr_extraction}

On CelebA-HQ dataset, we use string matching to extract attributes from the text. For example, the word ``bangs'' in the sentence indicates the ``bangs'' attribute. On CUB dataset, we extract attributes based on part-of-speech (POS) tags and dependency parsing implemented in spaCy~\cite{honnibal2017}. Concretely, given a text, we extract adjectives and nouns based on POS tags. Then, we leverage their dependency relations to extract the attributes. For example, in the text ``\textit{the bird has a yellow breast},'', ``yellow'' and ``breast'' has the adjectival modifier (amod) dependency relation, which indicates the ``yellow breast'' attribute. We also use other dependency relations to deal with sentences with more complex sentence structures. For example, in the text ``the bird has a brown and yellow breast,'' ``yellow'' and ``brown'' have the ``conjunct'' (conj) dependency relation, which indicates two attributes---``yellow breast'' and ``brown breast.''

\subsection{Pseudo-ground-truth Mask}
\label{subsec.supp.pseudo_gt_mask}

\begin{figure}
    \centering
    \includegraphics[width=\linewidth]{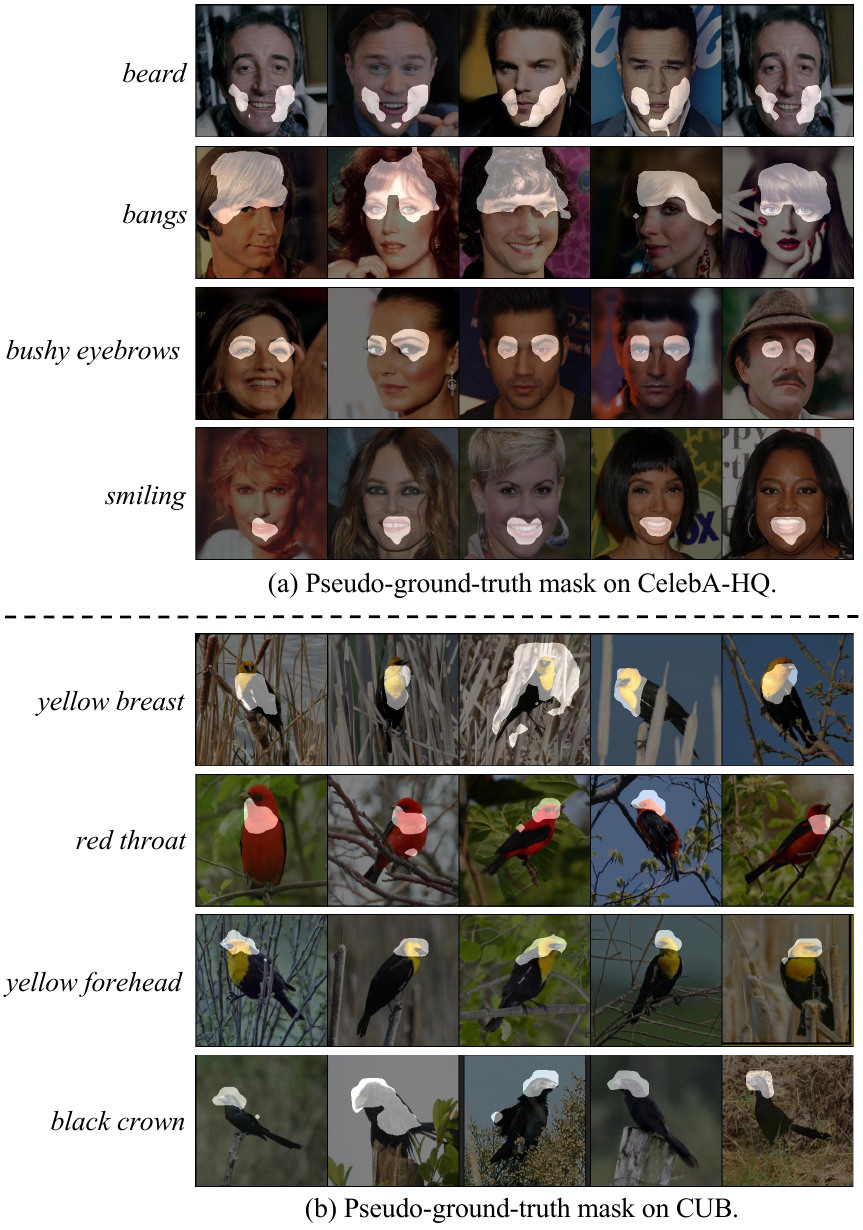}
    \caption{Pseudo-ground-truth masks generated by \cite{huang2020IEEEConf.Comput.Vis.PatternRecognit.CVPRb} on CelebA-HQ~\cite{karras2018Int.Conf.Learn.Represent.} and CUB~\cite{wah2011} datasets. The pseudo-ground-truth mask of the each attribute (\eg, \textit{beard}) is highlighted in white.}
    \label{fig.supp.pseudo_gt_mask}
\end{figure}

We use \cite{huang2020IEEEConf.Comput.Vis.PatternRecognit.CVPRb} as a weakly-supervised part segmentation network to obtain pseudo-ground-truth masks. The network is a classifier supervised by binary attribute labels extracted from text. In specific, since each image is paired with multiple texts, we use the union of attributes extracted from multiple texts as the image's attribute label. For example, if the image has two captions (1) ``\textit{the woman is smiling}'' and (2) ``\textit{the woman has blond hair},'' the attribute label for this image is (``\textit{woman}'', ``\textit{smiling},'' and ``\textit{blond hair}''). Based on these (image, binary attribute label) pairs, we train the network with binary cross-entropy loss. After training the network, we obtain an image's pseudo-ground-truth mask based on its attention map (Fig.~4 in \cite{huang2020IEEEConf.Comput.Vis.PatternRecognit.CVPRb}). We use Otsu method~\cite{otsu1979IEEETrans.Syst.ManCybern.} to threshold the attention map as the final pseudo mask ground-truth. Examples of pseudo-ground-truth mask are shown in \cref{fig.supp.pseudo_gt_mask}.

\subsection{Finetune CLIP}
\label{subsec.supp.finetune_clip}

We finetune the last few layers of CLIP. Specifically, we finetune the last visual resblock, ``ln\_post,'' ``proj'', the last text transformer resblock,  ``ln\_final'', ``text\_projection,'' and ``logit\_scale'' in CLIP. Following \cite{park2021Thirty-FifthConf.NeuralInf.Process.Syst.DatasetsBenchmarksTrackRound1}, we use AdamW~\cite{loshchilov2019Int.Conf.Learn.Represent.} optimizer and $5 \times 10^{-4}$ learning rate.

When finetuning CLIP for the \textit{CLIP-guided Contrastive Loss} (Eq.~1), the objective function for finetuning is contrastive loss defined in \cite{radford2021Int.Conf.Mach.Learn.}, where we use the (real image, text) pairs from the training split of the dataset for computing the contrastive loss.

As reported by \citet{zhang2021IEEEConf.Comput.Vis.PatternRecognit.CVPRa}, using the same model in training and testing can skew the R-Precision results. To alleviate this issue, for computing R-Precision results, we use a CLIP model that is different from the one used in training. We use the contrastive loss to finetune CLIP on the whole dataset (both training and testing splits), which is different from the CLIP used in training (finetuned on the training split only).

When finetuning CLIP for the \textit{Semantic Matching Loss} (\cref{eq.semantic_match_triplet}), the objective function for finetuning is binary cross-entropy loss. Concretely, the image's predicted probability of an attribute is computed by $\text{sigmoid}(\tau \cdot \cos(E_\text{CLIP}^\text{img}(\mathbf{I}), E_\text{CLIP}^\text{text}(\mathbf{t^a})))$. Here, $\mathbf{I}$ denotes an image. $\mathbf{t^a}$ denotes an attribute. $\tau$ is the ``logit\_scale'' parameter in CLIP optimized during finetuning. The predicted probability is used in binary cross-entropy to compute the loss.

\section{Ablation Studies of Text-to-Image}
\label{sec.supp.ablate_t2i}

We show more ablation studies results of text-to-image synthesis.

\paragraph{Results on CelebA-HQ} We show the ablation study results on CelebA-HQ dataset in \cref{tab.supp.ablate_t2i_celebahq}. The results are consistent with the ablation study results on CUB dataset in \cref{tab.ablate_t2i}, which further proves the effectiveness of each component of StyleT2I.

\begin{table}[t]
\centering
\begin{adjustbox}{width=\linewidth}
  \begin{tabular}{@{}lcc@{}}
\toprule
                    & R-Precision $\uparrow$    & FID $\downarrow$           \\ \midrule
w/o \textit{CLIP-guided Contrastive Loss} &  0.488       & \textbf{17.06} \\
w/o \textit{norm penalty}    & \textbf{0.736} & \underline{25.75}    \\
w/o \textit{Spatial Constraint}   & 0.607          & \textbf{17.45} \\
w/o \textit{Compositional Attribute Adjustment} & 0.594          & 17.59 \\
w/o finetune CLIP         & \underline{0.344}     & 17.79          \\
Full Model          & \textbf{0.625} & 17.46 \\ \bottomrule
\end{tabular}
\end{adjustbox}
\caption{Ablation study of StyleT2I on CelebA-HQ~\cite{karras2018Int.Conf.Learn.Represent.} dataset. Top-2 results are bolded and the worst results are underlined.}
\label{tab.supp.ablate_t2i_celebahq}
\end{table}

\paragraph{Threshold of norm penalty ($\theta$)} We conduct an ablation study on different threshold values ($\theta$) of norm penalty (\cref{eq.norm_penalty}). To better decide the threshold used for norm penalty, we compute the minimum (min), mean, and maximum (max) $\ell_2$ norm between two random latent codes sampled from $\mathcal{W}+$ space of StyleGAN (sampling from $\mathcal{W}+$ space is performed by feeding the sampled Gaussian noise to the ``Mapping Network'' in StyleGAN). We found that the minimum $\ell_2$ norm in StyleGAN trained on CelebA-HQ and CUB datasets are 8.2 and 8.9, respectively. Therefore, we choose $\theta = 8$ in our experiment to force the \textit{Text-to-Direction} and \textit{Attribute-to-Direction} modules find the direction with the smallest norm. As results shown in \cref{tab.supp.ablate_threshold_norm_penalty}, although larger $\theta$ can increase R-Precision results, it also renders worse image quality (larger FID values). Hence, using $\theta = 8$ strikes a nice balance between image-text balance and image quality.

\begin{table}[t]
\centering
\begin{adjustbox}{width=0.8\linewidth}
\begin{tabular}{@{}clll@{}}
\toprule
dataset                    & \multicolumn{1}{c}{threshold ($\theta$)} & \multicolumn{1}{c}{R-Precision} $\uparrow$ & \multicolumn{1}{c}{FID} $\downarrow$ \\ \midrule
\multirow{3}{*}{CelebA-HQ} & 8 (min)                   & 0.625                           & \textbf{17.46}                   \\
                           & 16 (mean)                 & \textbf{0.815}                           & 21.35                   \\
                           & 31 (max)                  & 0.801                           & 25.77                   \\ \midrule
\multirow{3}{*}{CUB}       & 8 (min)                   & 0.264                           & \textbf{20.53}                   \\
                           & 20 (mean)                 & \textbf{0.395}                           & 22.41                   \\
                           & 39 (max)                  & 0.375                           & 26.97                   \\ \bottomrule
\end{tabular}
\end{adjustbox}
\caption{Ablation study on the threshold of \textit{norm penalty} ($\theta$ in Eq.~2). Here, ``min'', ``mean'', and ``max'' stand for the minimum, average, and maximum $\ell_2$ norm of two randomly sampled latent codes of the pretrained StyleGAN. }
\label{tab.supp.ablate_threshold_norm_penalty}
\end{table}

\paragraph{Alternatives to norm penalty} We also tried other alternatives to improve image quality. One way is using the discriminator loss---making the synthesized image fool a discriminator. Another approach is using the perceptual loss to minimize the feature distance between the synthesized and real images. As the results shown in \cref{tab.supp.ablate_alternatives_to_norm_penalty}, our \textit{norm penalty} is the most effective way to ensure the image quality, while other approaches produce much higher FID values (\ie, worse image quality results).

\begin{table}[t]
\centering
\begin{adjustbox}{width=0.8\linewidth}
\begin{tabular}{@{}clc@{}}
\toprule
dataset                    & method for image quality & FID  $\downarrow$          \\ \midrule
\multirow{3}{*}{CelebA-HQ} & discriminator            & 32.83          \\
                           & perceptual loss          & 24.98          \\
                           & \textit{norm penalty} (\textbf{Ours})      & \textbf{17.46} \\ \midrule
\multirow{3}{*}{CUB}       & discriminator            & 26.25          \\
                           & perceptual loss          & 29.49          \\
                           & \textit{norm penalty} (\textbf{Ours})      & \textbf{20.53} \\ \bottomrule
\end{tabular}
\end{adjustbox}
\caption{Ablation study of different methods for improving image quality.}
\label{tab.supp.ablate_alternatives_to_norm_penalty}
\end{table}

\paragraph{Training Stage Regularization} We create an alternative to \textit{Compositional Attribute Adjustment}---``Training Stage Regularization.'' While our \textit{Compositional Attribute Adjustment} adjusts the sentence direction during the inference stage, ``Training Stage Regularization'' maximizes the cosine similarity between the sentence direction and attribute directions, \ie, $\max \sum_i \cos( \mathbf{s}, \mathbf{a}_i )$, which is added as an additional loss to Eq.~3 to regularize the \textit{Text-to-Direction} module during the training stage. The results comparing the ``Training Stage Regularization'' and \textit{Compositional Attribute Adjustment} are shown in \cref{tab.supp.training_stage_regularization}. Two methods achieve similar FID results. However, our \textit{Compositional Attribute Adjustment} achieves better R-Precision results than ``Training Stage Regularization.'' We believe the reason is that regularizing during the training stage only helps for seen attribute compositions in the training set, which cannot ensure the correct attribute prediction during the inference stage. Therefore, our proposed \textit{Compositional Attribute Adjustment} can better improve the image-text alignment by
adjusting the results during the inference stage for text with unseen attribute compositions.

\begin{figure}
    \centering
    \includegraphics[width=\linewidth]{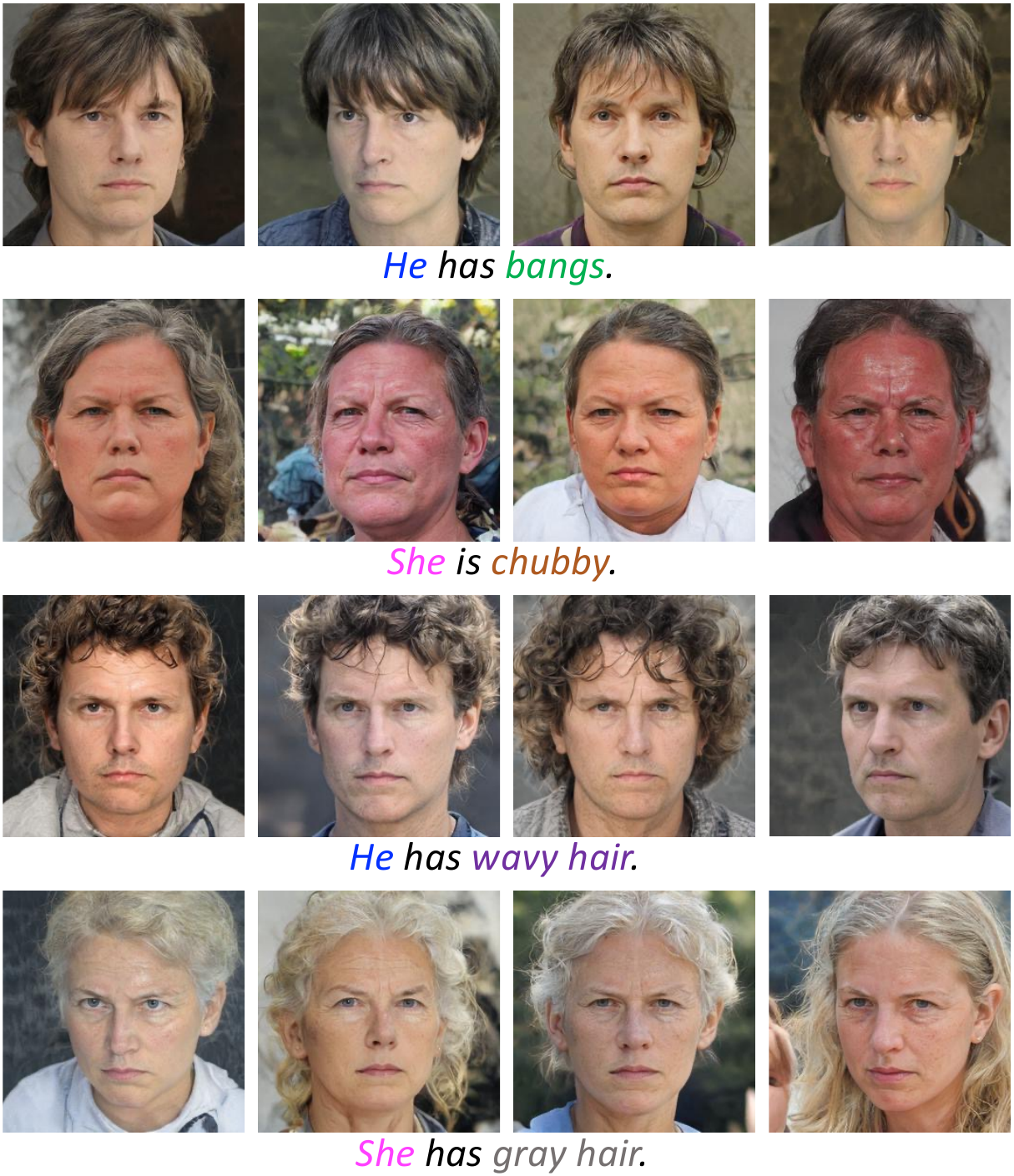}
    \caption{Diverse results when sampling four different $\mathbf{z}$.}
    \label{fig.supp.random_z}
\end{figure}

\paragraph{Different $\mathbf{z}$} We sample three different $\mathbf{z}$ for each text to compute the standard deviation of R-Precision, which is 0.008, proving that $\mathbf{z}$ does not have a significant effect on the image-text alignment. The synthesized images of the same text in various $\mathbf{z}$ in \cref{fig.supp.random_z}, proving the diversity of the synthesis results.

\begin{table}[]
\centering
\begin{tabular}{lll}
\hline
Method          & R-Precision $\uparrow$  & FID $\downarrow$   \\ \hline
ControlGAN      & 0.498          & 17.36          \\
DAE-GAN         & 0.546          & 19.24          \\
TediGAN-A       & 0.026          & \textbf{12.92} \\
TediGAN-B       & 0.354          & 14.19          \\
StyleT2I (\textbf{Ours}) & \textbf{0.635} & 15.60          \\ \hline
\end{tabular}
\caption{Results on CelebA-HQ's standard split.}
\label{tab.supp.celebahq_standard_split}
\end{table}

\paragraph{Results on CelebA-HQ's standard split}
We also show the results on the CelebA-HQ's standard testing split, \ie, not the test split that we created for the evaluation of compositionality (\cref{subsec.exp_setup}), in \cref{tab.supp.celebahq_standard_split}. Most of the results are better than the results on the new split (\cref{tab.main_t2i}) because of the overlap between train and test splits that allows the models to cheat.

\begin{table}[t]
\centering
\begin{adjustbox}{width=\linewidth}
  \begin{tabular}{@{}clcc@{}}
\toprule
dataset                    & \multicolumn{1}{c}{method}         & R-Precision $\uparrow$  & FID $\downarrow$ \\ \midrule
\multirow{2}{*}{CelebA-HQ} & Training Stage Regularization      & 0.604                           & 17.56                   \\
                           & \textit{Compositional Attribute Adjustment} & \textbf{0.625}                           & \textbf{17.46}                   \\ \midrule
\multirow{2}{*}{CUB}       & Training Stage Regularization      & 0.256                           & \textbf{19.48}    \\
                           & \textit{Compositional Attribute Adjustment} & \textbf{0.264}                           & 20.53                   \\ \bottomrule
\end{tabular}
\end{adjustbox}
\caption{Ablation study of \textit{Compositional Attribute Adjustment}. ``Training Stage Regularization'' stands for using attribute directions to supervise the the sentence direction during the training stage, which can be regarded as an alternative method to \textit{Compositional Attribute Adjustment} that uses attribute directions to adjust sentence direction during the inference stage.}
\label{tab.supp.training_stage_regularization}
\end{table}

\section{Ablation Studies of Identifying Attribute Directions}
\label{sec.supp.ablate_w2d}

We further conduct more ablation studies of identifying attribute directions on CelebA-HQ dataset. To evaluate the identified attribute directions, we train a ResNet-18 classifier with the ground-truth attribute labels (\ie, not the labels extracted from text) as the attribute classifier. We use this attribute classifier to evaluate the synthesized positive and negative images generated from \textit{Attribute-to-Direction} module (\cref{fig.triplet}). For the positive image, its attribute ground-truth is positive. For the negative image, its attribute ground-truth is negative. We compute \textit{Attribute Accuracy} based on the attribute classifier's prediction and ground-truth. Higher \textit{Attribute Accuracy} indicates a more accurate attribute direction.

\paragraph{Spatial Constraint}

\begin{table}[]
\centering
\begin{tabular}{@{}rc@{}}
\toprule
& Attribute Accuracy $\uparrow$ \\ \midrule
w/o \textit{Spatial Constraint} & 0.827 \\
w/ \textit{Spatial Constraint}         & \textbf{0.871} \\ \bottomrule
\end{tabular}
\caption{Ablation study of \textit{Spatial Constraint} for identifying attribute directions on CelebA-HQ dataset.}
\label{tab.supp.ablate_spatial_constraint_for_attr_to_direction}
\end{table}

The results of the ablation study on \textit{Spatial Constraint} are shown in \cref{tab.supp.ablate_spatial_constraint_for_attr_to_direction}, which proves that \textit{Spatial Constraint} can help the \textit{Attribute-to-Direction} module find more accurate attribute directions by leveraging the intended region from pseudo-ground-truth mask.

\paragraph{Margin of Semantic Matching Loss ($\alpha$)}

We conduct the ablation study on the margin ($\alpha$) of \textit{Semantic Matching Loss} (\cref{eq.semantic_match_triplet}). The results in \cref{tab.supp.ablate_margin} show that the results are converged when $\alpha \geq 1$. We choose $\alpha = 1$ in the main experiments.

\begin{table}[t]
\centering
\begin{tabular}{@{}cc@{}}
\toprule
margin & Attribute Accuracy \\ \midrule
0.1    & 0.577    \\
0.5    & 0.761    \\
1      & 0.871    \\
5      & 0.881    \\
10     & 0.875    \\
20     & 0.873    \\ \bottomrule
\end{tabular}
\caption{Ablation study on the margin ($\alpha$) of \textit{Semantic Matching Loss} on CelebA-HQ dataset. The accuracy results are not sensitive to the value of margin when $\alpha \geq 1$.}
\label{tab.supp.ablate_margin}
\end{table}

\paragraph{Alternative to Spatial Constraint}

An alternative approach to improve disentanglement among different attributes is encouraging different attribute directions to be orthogonal with each other in the latent space~\cite{shen2020IEEEConf.Comput.Vis.PatternRecognit.CVPR}. Therefore, we create an alternative approach by minimizing $\sum_i \sum_j \frac{\mathbf{a}_i}{||\mathbf{a}_i||_2}^T \frac{\mathbf{a}_j}{||\mathbf{a}_j||_2}$ when training the \textit{Attribute-to-Direction} module. The results in \cref{tab.supp.ablate_orth_penalty} show that this alternative approach hurts the accuracy performance compared with only using the \textit{Semantic Matching Loss}. In contrast, our \textit{Spatial Constraint} can greatly improve the accuracy results.

\begin{figure*}
    \centering
    \includegraphics[width=0.8\linewidth]{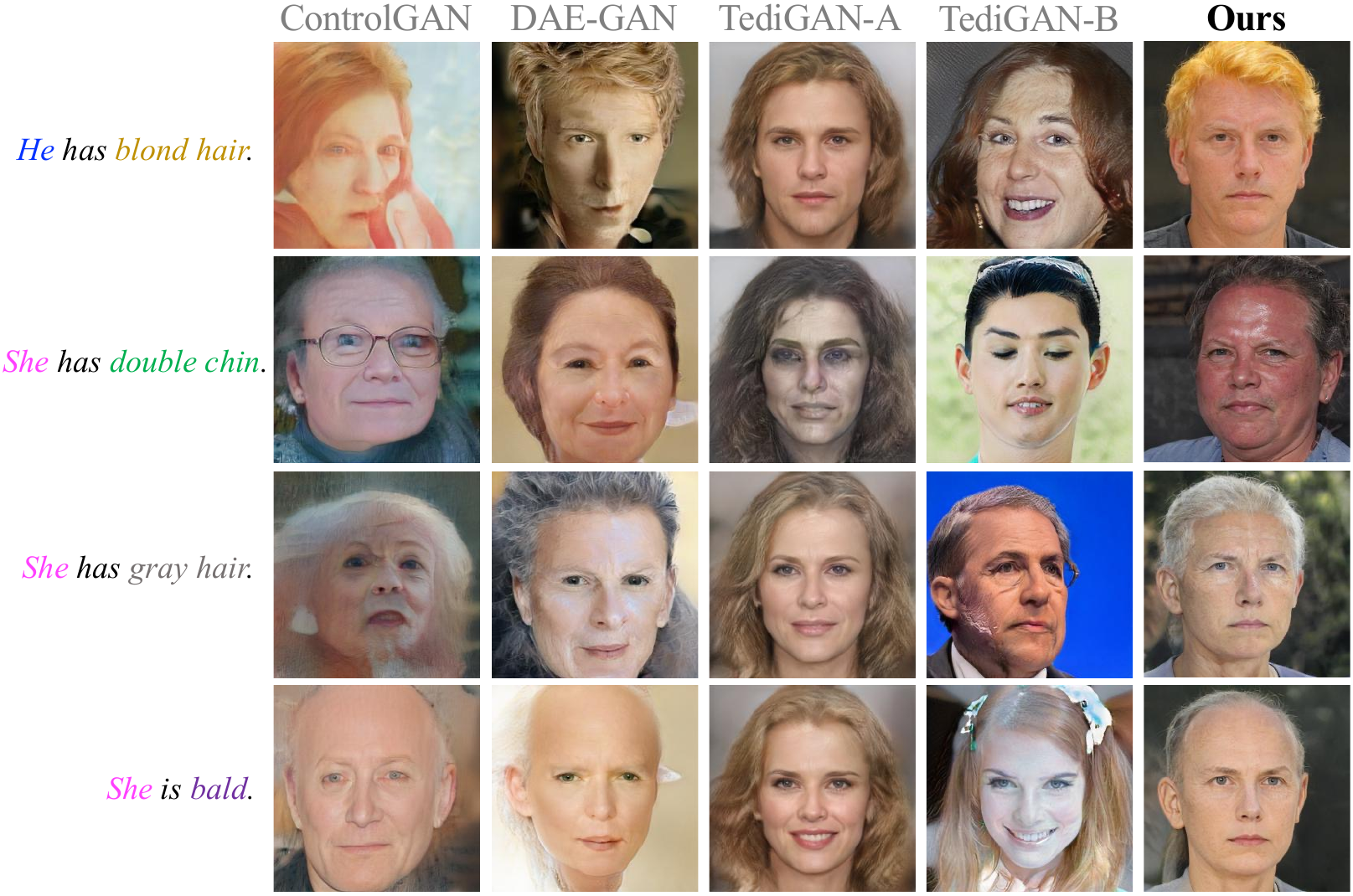}
    \caption{More examples of synthesis results where the input text decribes underrepresented compositions of attribute on CelebA-HQ dataset.}
    \label{fig.supp.underrepresented}
\end{figure*}

\begin{table}[t]
\centering
\begin{tabular}{@{}lc@{}}
\toprule
                   & Attribute Accuracy       \\ \midrule
\textit{Semantic Matching Loss} only & 0.827 \\
w/ $\min \sum_i \sum_j \frac{\mathbf{a}_i}{||\mathbf{a}_i||_2}^T \frac{\mathbf{a}_j}{||\mathbf{a}_j||_2} $      & 0.809          \\
w/ \textit{Spatial Constraint} & \textbf{0.871} \\ \bottomrule
\end{tabular}
\caption{Comparison between \textit{Spatial Constraint} and an alternative approach $\min \sum_i \sum_j \frac{\mathbf{a}_i}{||\mathbf{a}_i||_2}^T \frac{\mathbf{a}_j}{||\mathbf{a}_j||_2} $ for disentanglement on CelebA-HQ dataset. \textit{Spatial Constraint} achieves better results.}
\label{tab.supp.ablate_orth_penalty}
\end{table}

\paragraph{Alternative to Semantic Matching Loss---Contrastive Loss} Since the \textit{Text-to-Direction} module and \textit{Attribute-to-Direction} module share some similarity, one may wonder if it is feasible to use the contrastive loss to train the \textit{Attribute-to-Direction}. To this end, we adapt our \textit{CLIP-guided Contrastive Loss} for \textit{Attribute-to-Direction} module by replacing the text input with attribute input, which attracts the embeddings of paired synthesized image and attribute and repels the embeddings of mismatched pairs.

The results of comparing this alternative method and \textit{Semantic Matching Loss} are shown in \cref{tab.supp.ablate_contrastive_loss_a2d}. The contrastive loss achieves poorer performance for identifying attribute directions. The reason is that we should not repel the embeddings mismatched (image, attribute) pairs. For example, we should not repel the embedding of an ``\textit{smiling}'' image against ``\textit{man}'' attribute when the random latent code $\mathbf{z}$ can be used to synthesize a male face image. Therefore, our \textit{Semantic Matching Loss} can identify the attribute directions better since it does not repel the embeddings of mismatched (image, attribute) pairs.

\begin{table}[]
\centering
\begin{adjustbox}{width=\linewidth}
  \begin{tabular}{@{}lc@{}}
\toprule
                       & Attribute Accuracy       \\ \midrule
Contrastive Loss + \textit{Spatial Constraint}      & 0.669 \\
\textit{Semantic Matching Loss} + \textit{Spatial Constraint} & \textbf{0.871}          \\ \bottomrule
\end{tabular}
\end{adjustbox}
\caption{Ablation study of \textit{Semantic Matching Loss} for identifying attribute directions on CelebA-HQ dataset.}
\label{tab.supp.ablate_contrastive_loss_a2d}
\end{table}

\paragraph{Local Direction vs. Global Direction} Our \textit{Attribute-to-Direction} module predicts the attribute direction conditioned on both input attribute and random latent code $\mathbf{z}$. One may wonder if conditioning on the random latent code is necessary. Following the terms defined by \citet{zhuang2021Int.Conf.Learn.Represent.}, we call the attribute direction conditioned on the random latent code as ``local direction,'' and we name the attribute direction only conditioned on the attribute (\ie, not conditioned on random latent code) as ``global direction.'' The results comparing local direction and global direction are shown in \cref{tab.supp.ablate_global_local_direction}. The global direction, which predicts a single direction for an attribute globally, achieves poor attribute accuracy results. In contrast, our local direction method, which takes the random latent code into the consideration, can more accurately predict the attribute direction.

\begin{table}[t]
\centering
\begin{tabular}{@{}lc@{}}
\toprule
                 & Attribute Accuracy       \\ \midrule
global direction & 0.764          \\
local direction (\textbf{Ours})  & \textbf{0.871} \\ \bottomrule
\end{tabular}
\caption{Ablation study of global direction vs. local direction for identifying attribute directions on CelebA-HQ dataset.}
\label{tab.supp.ablate_global_local_direction}
\end{table}

\begin{figure*}
    \centering
    \includegraphics[width=0.8\linewidth]{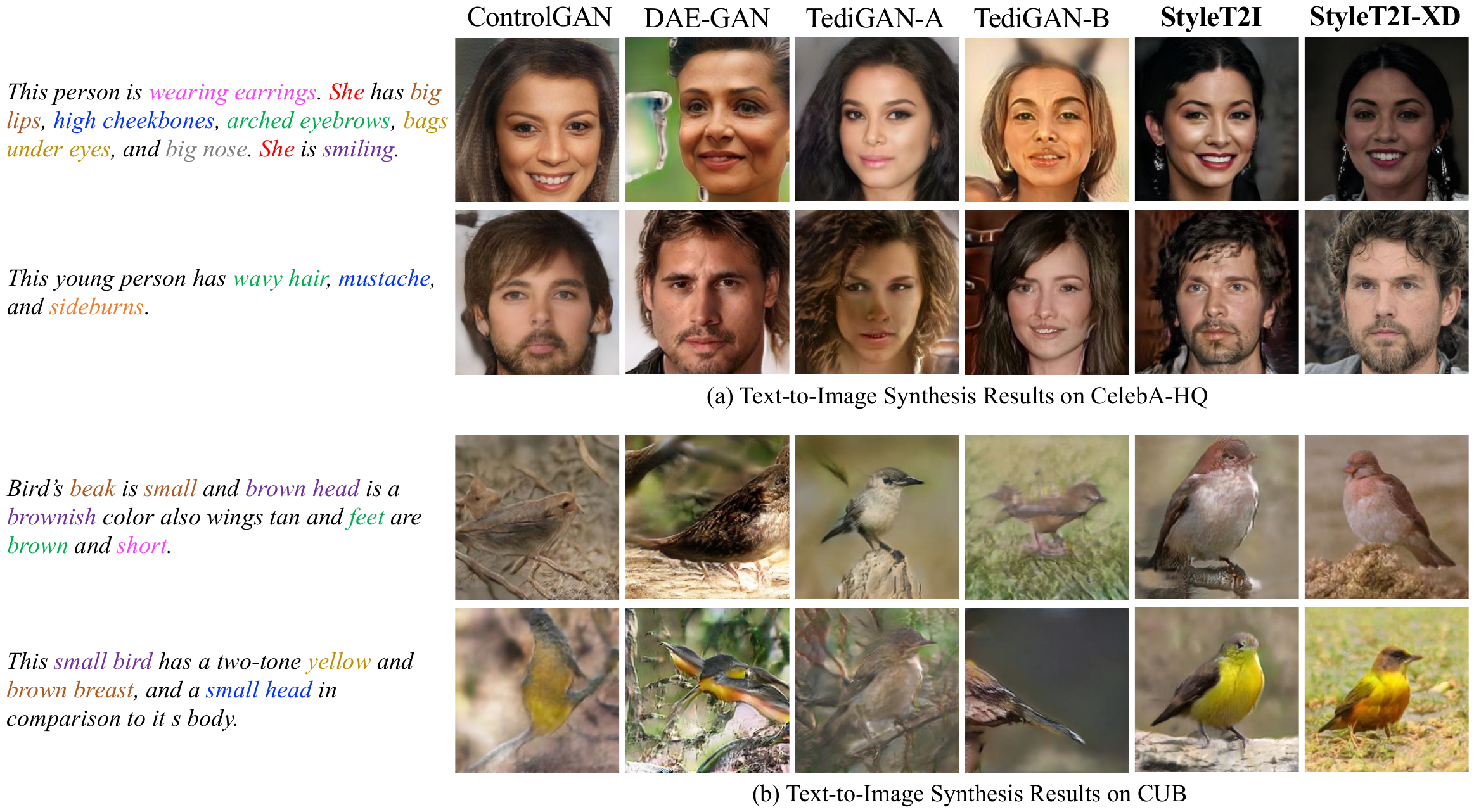}
    \caption{More examples of text-to-image synthesis results.}
    \label{fig.supp.t2i_more_examples}
\end{figure*}

\section{More Qualitative Results}
\label{sec.supp.more_qualitative}

\paragraph{Underrepresented Compositions}

More examples of synthesis results where the input texts describe underrepresented compositions of attributes are shown in \cref{fig.supp.underrepresented}. Our method can more accurately synthesize the image for underrepresented attribute compositions with high image fidelity.

\paragraph{Text-to-Image Results}

More examples of text-to-image synthesis results are shown in \cref{fig.supp.t2i_more_examples}. Our method can synthesize images conditioned on the text describing unseen attribute compositions with better image-text alignment and higher image quality.

\paragraph{Norm Penalty}

\begin{figure}[t]
    \centering
    \includegraphics[width=\linewidth]{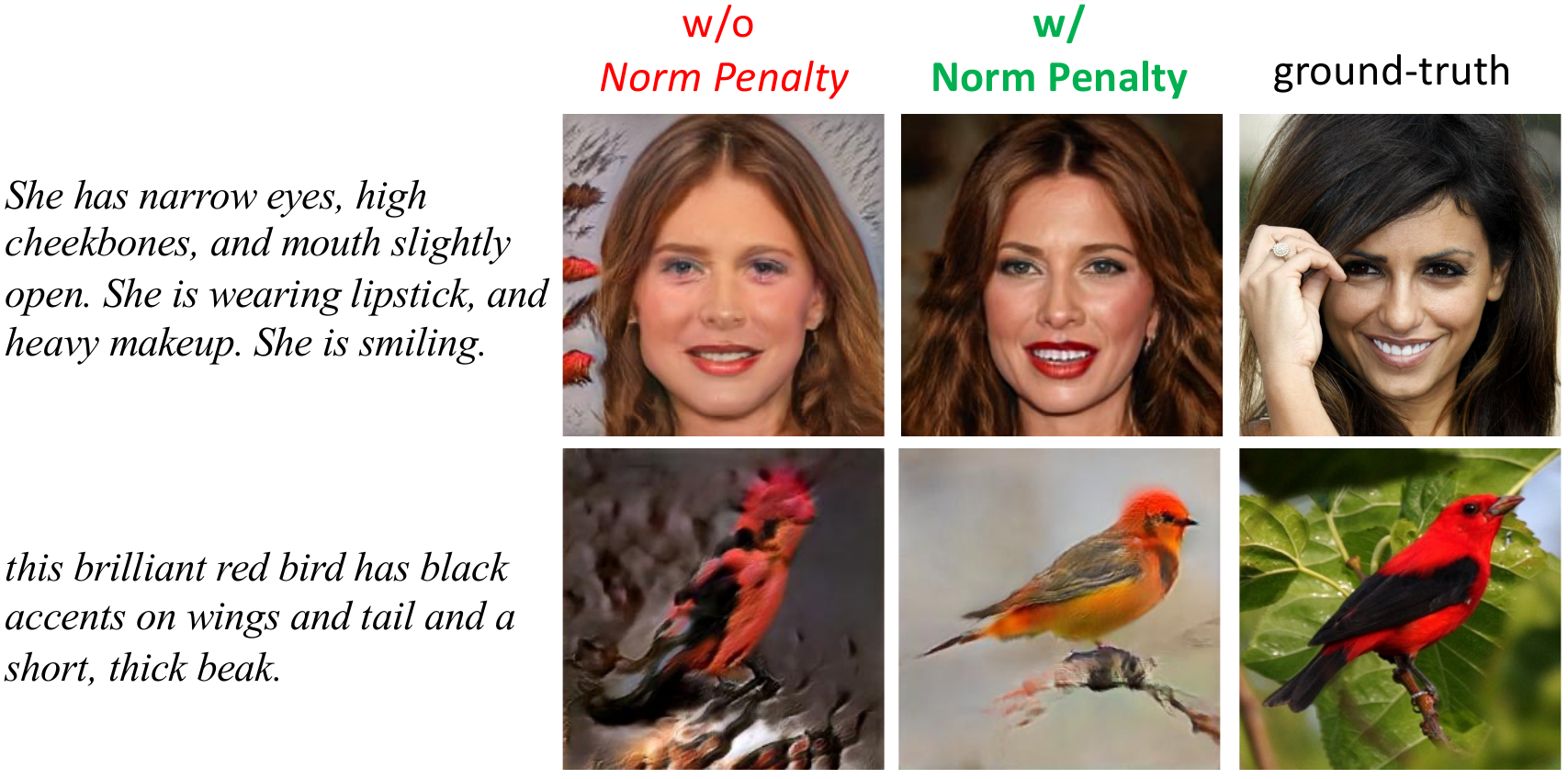}
    \caption{More examples of the ablation study on \textit{norm penalty}.}
    \label{fig.supp.ablate_norm_penalty}
\end{figure}

More examples of the ablation study on \textit{norm penalty} are shown in \cref{fig.supp.ablate_norm_penalty}, which proves that \textit{norm penalty} can effectively improve the image quality.

\begin{figure}[t]
    \centering
    \includegraphics[width=\linewidth]{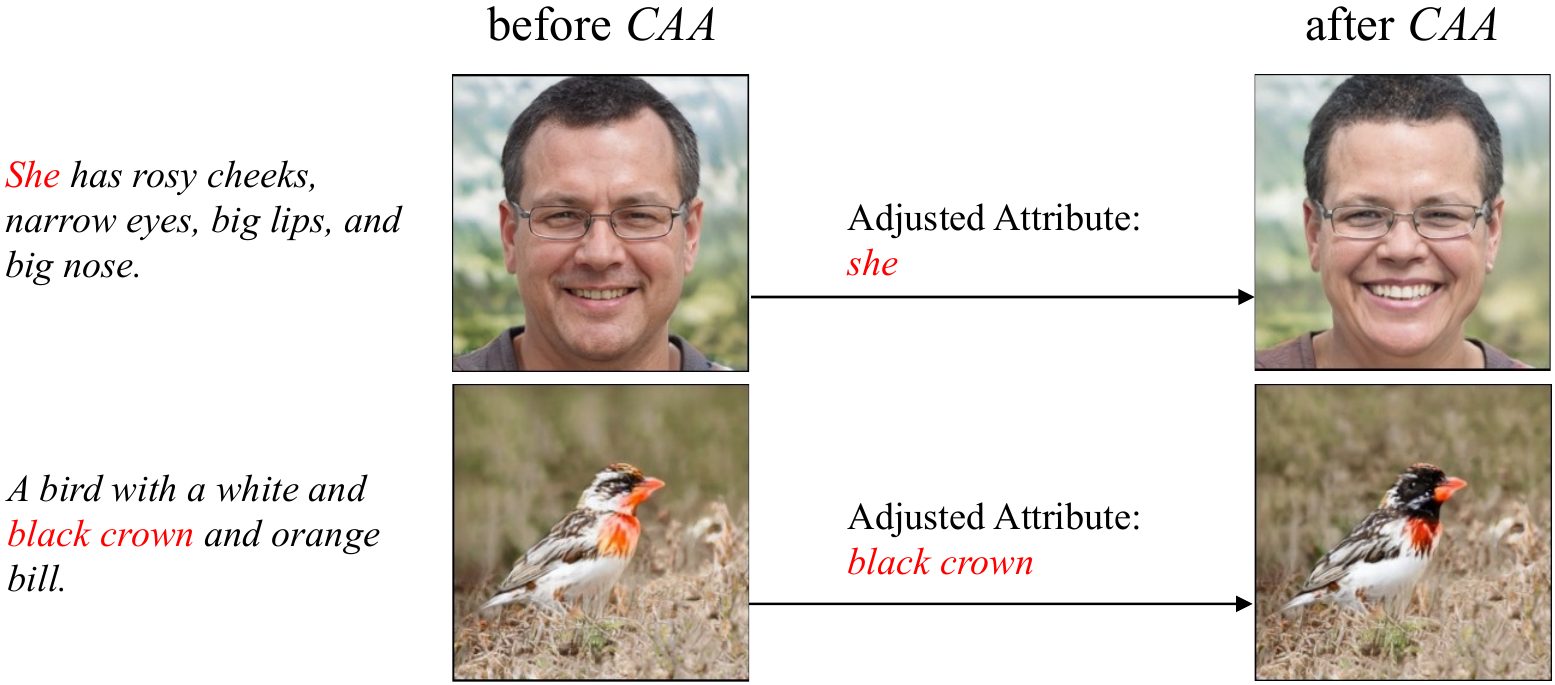}
    \caption{More examples of the ablation study on \textit{Compositional Attribute Adjustment} (\textit{CAA}).}
    \label{fig.supp.ablate_caa}
\end{figure}

\paragraph{Compositional Attribute Adjustment}

More examples of the ablation study on \textit{Compositional Attribute Adjustment} (\textit{CAA}) are shown in \cref{fig.supp.ablate_caa}, which demonstrates that \textit{CAA} can automatically identify the wrong attribute predictions and effectively correct them during the inference stage to improve the compositionality.

\begin{figure*}[t]
    \centering
    \includegraphics[width=0.8\linewidth]{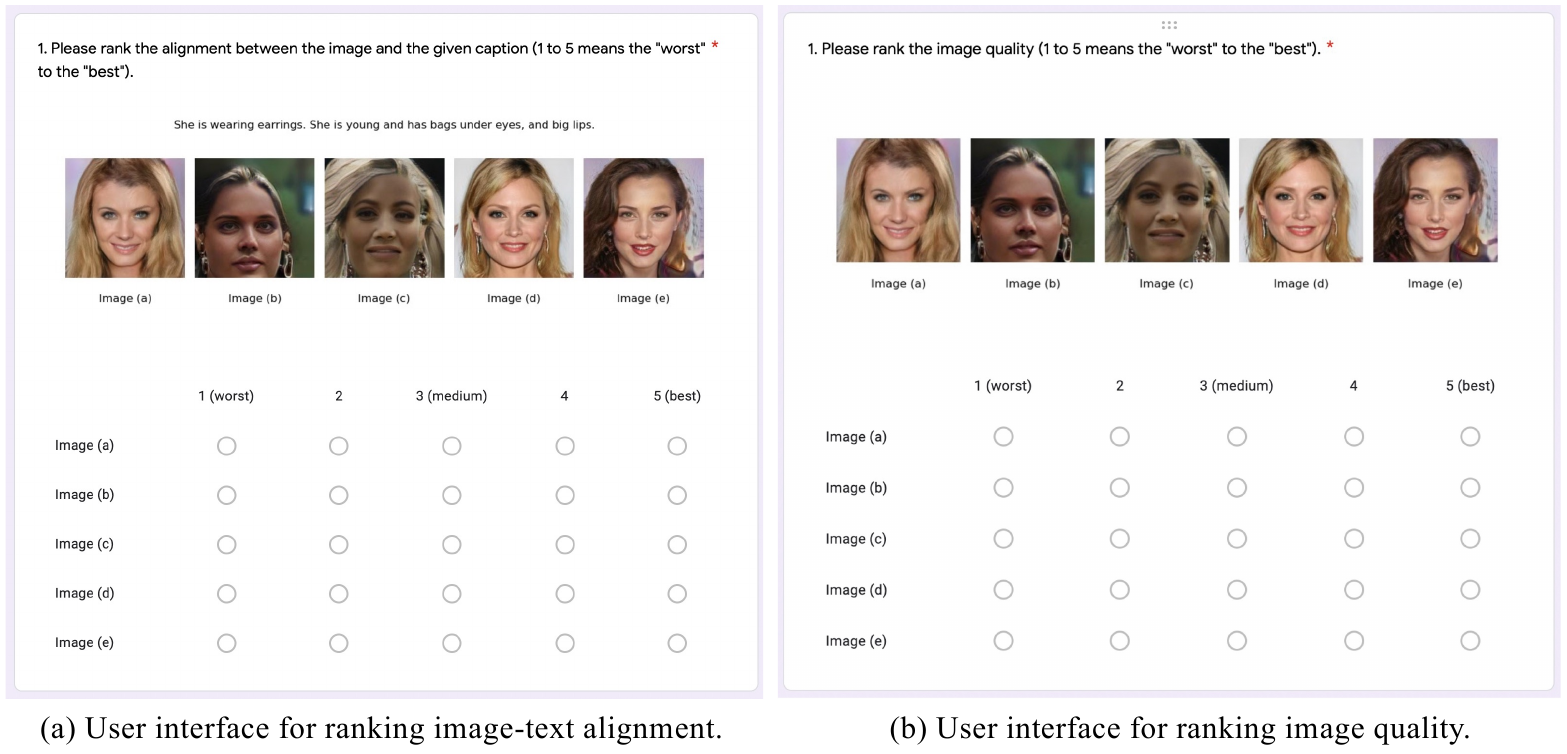}
    \caption{User interface for user study.}
    \label{fig.supp.user_interface}
\end{figure*}

\section{User Study}
\label{sec.supp.user_study}

On each dataset, we randomly sample 20 sentences from the testing split to synthesize the images for the user study. We invite 12 participants to evaluate the image-text alignment and the image quality.

We request the participants to read a guideline before conducting the user study. For evaluating the image-text alignment on face images, our guideline clarifies that the words like ``woman,'' ``man,'' ``she,'' ``he'' denote the visually perceived gender, which does not imply one's real gender identity. Since participants may not be familiar with some terms in the birds image domain, we provide Fig.~2 in \cite{wah2011}, an illustration of fine-grained bird part names (\eg, nape), in the guideline of the user study to help participants better understand the text.

We use Google Form to collect the user study results.
The user interface for the user study is shown in \cref{fig.supp.user_interface}. The method names are not shown in the user interface. In each question, the order of images generated from different methods is shuffled.

The user study in this paper follows the research protocol, whose master study received the exempt determination from Institutional Review Board (IRB).

\section{Discussion}
\label{sec.supp.discussion}

\subsection{Limitations and Future Research Directions}
We honestly list some limitations of our work and discuss some promising future research directions.

First, our attribute extraction approach (\cref{subsec.attr_extraction}) is limited by assuming that adjectives and nouns in the text can imply the attribute, which cannot be generalized to texts describing more complex relations in the image. For example, the text ``\textit{the earring on the left is bigger than the earring on the right},'' describes a relative relation (\eg, ``\textit{bigger}''), which cannot be expressed as an attribute.

Second, based on StyleGAN, StyleT2I focuses on synthesizing find-grained images in face and bird domains, where StyleGAN has shown a great capability of synthesizing high-fidelity images. However, our initial experiment finds that StyleGAN cannot synthesize high-quality complex scene images from MS-COCO~\cite{lin2014Eur.Conf.Comput.Vis.ECCV,chen2015ArXiv150400325Cs} dataset, which limits our method to focus on fine-grained single-object image domains, \eg, faces and birds. Future works can study how to leverage pretrained scene image generators (\eg, SPADE~\cite{park2019IEEEConf.Comput.Vis.PatternRecognit.CVPR}) to perform text-to-image synthesis.

Third, in terms of \textit{Spatial Constraint}, the pseudo-ground-truth masks for some images are not accurate, which introduces label noises for \textit{Spatial Constraint}. Future work can leverage some recent semi-supervised methods to obtain the pseudo-ground-truth mask for \textit{Spatial Constraint}. For example, by only annotating a few images, \cite{zhang2021IEEEConf.Comput.Vis.PatternRecognit.CVPR} uses StyleGAN to synthesize high-quality images with pseudo-ground-truth masks, which can be used as an alternative to the weakly-supervised method~\cite{huang2020IEEEConf.Comput.Vis.PatternRecognit.CVPRb} used in this work.

\subsection{Potential Negative Societal Impacts}

Since StyleT2I can synthesize high-fidelity images, a malicious agent may use our model as a deepfake technology for unintended usage. To mitigate this issue, we ask the users to agree to the ethics terms when releasing the model.
Overall, StyleT2I improves the compositionality of text-to-image synthesis, which can better synthesize images for text containing underrepresented attribute compositions, \eg, ``\textit{he is wearing lipstick}.'' Therefore, we believe that StyleT2I contributes to reducing the negative societal impact compared with previous text-to-image synthesis methods.

\end{document}